%% file: main.tex
\pdfoutput=1

\documentclass[11pt]{article}

\usepackage{acl}

\usepackage{times}
\usepackage{latexsym}

\usepackage[T1]{fontenc}

\usepackage[utf8]{inputenc}

\usepackage{microtype}

\usepackage{inconsolata}

\usepackage{settings}

\title{Metaphor Understanding Challenge Dataset for LLMs}

\author{Xiaoyu Tong$^{\heartsuit}$ \and Rochelle Choenni$^{\heartsuit}$ \and Martha Lewis$^{\diamondsuit}$ \and Ekaterina Shutova$^{\heartsuit}$ \\
$^{\heartsuit}$ILLC, University of Amsterdam, the Netherlands \\
$^{\diamondsuit}$Dept. of Engineering Mathematics, University of Bristol, UK \\
\texttt{\{x.tong,r.m.v.k.choenni,e.shutova\}@uva.nl} \\
\texttt{marthalewis@santafe.edu}}

\begin{document}
\maketitle
\begin{abstract}
\input{body/0_abstract}
\end{abstract}

\section{Introduction}
\label{sec:intro}
\input{body/1_intro}

\section{Related work}
\label{sec:related}

\input{body/2_related_work}

\section{Data collection: metaphor samples}

\input{body/3_data_collection}

\section{Annotation of apt paraphrases}
\label{sec:ann_apt}
\input{body/4_ann_apt}

\section{Annotation of inapt paraphrases}
\label{sec:ann_inapt}

\input{body/5_ann_inapt}

\section{Data analysis}
\input{body/6_data_analysis}

\section{Model evaluation}
\input{body/7_model_eval}

\section{Discussion}
\input{body/8_discussion}

\section{Conclusion}

We release a dataset of manually created apt and inapt paraphrases for metaphorical sentences and present two metaphor understanding tasks, which we demonstrate to be challenging for current LLMs.
The errors the models make in the paraphrase generation task indicate various levels of misunderstanding of the metaphors.
In the paraphrase judgement task, the models' accuracy was lower than the random baseline in the majority of the cases; a closer look at their errors reveals that the models had difficulty in detecting the inaptness of the inapt paraphrases.
The experiments also show that the models performed better when being instructed to focus on the metaphorical word, and that genre and the POS and novelty of the metaphorical word are all potential factors that affect model performance.

\section{Limitations}

We designed the metaphor understanding tasks to be representative of a lexical substitution task:
The metaphorical word is the only difference between a reference sentence and a candidate paraphrase.
This setup makes it possible to examine whether LLMs indeed perform metaphor interpretation or resort to lexical similarity when they encounter metaphorically used words.
At the same time, however, it also means that 
the models' understanding of multi-word metaphors and direct metaphors (e.g., similies) could not be tested using our dataset.

Another limitation is that the tested models are not the latest LLMs---Llama2 and GPT-4 were released shortly after the completion of our study.

\section{Ethics statement}

We abide by the ACL Code of Ethics.
The metaphor resources used in this study are publicly available.
The crowdsourcing task was approved by an ethics committee.
The crowd workers received fair payment (9 GBP per hour),
and no personal information was collected or stored in our database.

The metaphor samples in our dataset come from excerpts of natural discourse.
They may therefore involve bias, taboo, violence, or other aspects of everyday language use that could be considered negative (we also pointed this out to the crowd workers before they gave their consent to participate, as presented in Appendix~\ref{appx:pptask}).
Nonetheless, these are integral parts of language use, and should be properly understood by NLP systems, which is precisely what this paper aims at.

\section*{Acknowledgements}

This study is funded by the Faculty of Engineering at the University of Bristol, through the faculty's Pump-Priming Fund.
We thank anonymous reviewers for their constructive feedback.

\bibliography{anthology,custom}

\appendix

\section{Previous metaphor understanding datasets and tasks}
\label{appx:prev_datasets}
\input{body/a2_related_datasets}

\section{Crowdsourcing task}
\label{appx:pptask}

\input{body/a4_pptask}

\section{Inapt paraphrase annotation}
\label{appx:inapt}
\input{body/a5_inapt}

\section{Novelty distribution of MUNCH metaphor samples}
\label{appx:novelty}

\input{body/a6_novelty}

\section{Model evaluation details}
\label{appx:prompts}
\input{body/a7_prompts}

\section{Error analysis details}
\label{appx:error}
\input{body/a8_error}

\end{document}

%% file: body/0_abstract.tex
Metaphors in natural language are a reflection of fundamental cognitive processes such as analogical reasoning and categorisation,
and are deeply rooted in everyday communication.
Metaphor understanding is therefore an essential task for large language models (LLMs).
We release the \textbf{M}etaphor \textbf{Un}derstanding \textbf{Ch}allenge Dataset (MUNCH), designed to evaluate the metaphor understanding capabilities of LLMs.
The dataset provides over 10k paraphrases for sentences containing metaphor use, as well as 1.5k instances containing inapt paraphrases. The inapt paraphrases were carefully selected to serve as control to determine whether the model indeed performs full metaphor interpretation or rather resorts to lexical similarity.
All apt and inapt paraphrases were manually annotated.
The metaphorical sentences cover natural metaphor uses across 4 genres (academic, news, fiction, and conversation), and they exhibit different levels of novelty.
Experiments with LLaMA and GPT-3.5 demonstrate that MUNCH presents a challenging task for LLMs.
The dataset is freely accessible at
\url{https://github.com/xiaoyuisrain/metaphor-understanding-challenge}.

%% file: body/1_intro.tex
Large language models (LLMs), such as BERT \citep{devlin-etal-2019-bert}, GPT-3 \citep{brown-etal-2020-gpt3} and LLaMA \citep{touvron-etal-2023-llama}, have become a common paradigm in natural language processing (NLP).
Several benchmarks have been proposed to investigate the capabilities of LLMs \citep{srivastava-2022-poirot,liang-etal-2022-holistic,hendrycks-etal-2021-measuring}; and comprehensive analyses have been conducted, evaluating their performance on a range of NLU tasks \citep{zhong-etal-2023-chatgpt,qin-etal-2023-chatgpt,kocon-etal-2023-chatgpt,ye-etal-2023-comprehensive,bang-etal-2023-multitask}.
The community has extensively examined LLM performance on question answering, summarisation, sentiment analysis, natural language inference; a few studies have also shed light on LLMs' analogical reasoning capabilities \citep{czinczoll-etal-2022-scientific,webb-etal-2023-emergent}. However, the ability of LLMs to comprehend metaphor---a fundamental linguistic and cognitive tool---is still poorly understood.

Metaphors are linguistic expressions based on conceptual mappings between a target and a source domain \citep{lakoff-johnson-1980-metaphors}.
The verb phrase \lingform{to \underline{stir} excitement}, for example, is based on the conceptual metaphor \concept{feeling is liquid}, with \concept{feeling} (excitement) being the target domain and \concept{liquid} (something that can be stirred) the source domain.
The metaphor compares \concept{feeling} with \concept{liquid}, introducing vividness into the description of an otherwise intangible emotional impact. Such cross-domain mappings are sets of systematic ontological correspondences, mapping concepts and their relational structure across distinct domains. Performing this mapping is an essential part of reasoning involved in the interpretation of metaphorical language \cite{lakoff-2014-mapping,grady-etal-1999-blending,gentner-markman-1997-structure}.

\begin{figure}[!t]
    \centering
    \includegraphics[width=0.45\textwidth]{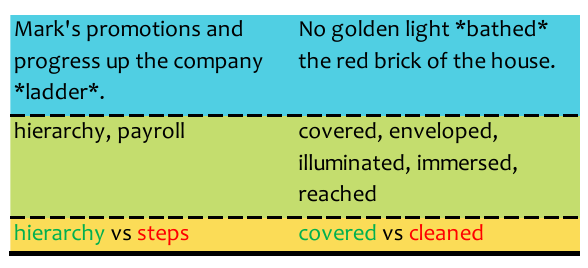}
    \caption{\label{fig:intro_dataset}
    MUNCH dataset samples. Each \colorbox{SkyBlue}{metaphor sample}
    has a $\ast$highlighted$\ast$ word that is metaphorically used, and is accompanied by up to 5 \colorbox{SpringGreen}{crowdsourced paraphrases}: Substituting the highlighted word with one of the provided words should result in an apt paraphrase.
    For a selection of metaphor samples, we also provide \colorbox{Goldenrod}{expert annotation}: a pair of \textcolor{Green}{correct} and \textcolor{red}{incorrect} substitution words.}
\end{figure}

\begin{figure}[!t]
    \centering
    \includegraphics[scale=0.50]{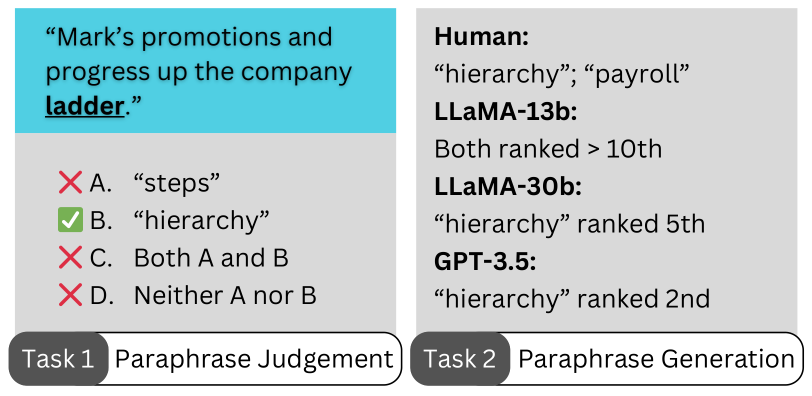}
    \caption{
     Two tasks for MUNCH: 
    Given a sentence containing a metaphorically used word, a model is prompted to 1) select correct paraphrases from two given candidates (Paraphrase Judgement), and 2) paraphrase the sentence by replacing the highlighted metaphorically used word (Paraphrase Generation).}
    \label{fig:intro_tasks}
\end{figure}

Humans use metaphors so naturally and frequently that they largely fly under our radar.
In one of the largest metaphor corpora annotated by linguists, the VU Amsterdam Metaphor Corpus (VUA; \citealp{steen-etal-2010-mipvu}), every 8th word is metaphorical,  
as averaged over four different genres, including academic and conversation.
LLMs, therefore, require the ability to comprehend 
metaphor in order to have a full command of language.
As such, metaphor understanding is an essential task for evaluating the capabilities of LLMs.

Several corpora have been created that contain metaphor annotations at either word or sentence level. These include the VUA corpus \cite{steen-etal-2010-mipvu}, the LCC metaphor datasets \cite{mohler-etal-2016-introducing} and the metaphor-emotion dataset of \citet{mohammad-etal-2016-metaphor}, among others. These datasets have been widely used to develop and evaluate automated metaphor identification systems (see \citet{tong-etal-2021-recent} for a survey), but they do not contain information of how the annotated metaphors are interpreted. On the other hand, several works developed datasets with a focus on interpretation, typically casting the problem as a paraphrasing task \cite{shutova-2010-automatic,bizzoni-lappin-2018-predicting,joseph-etal-2023-newsmet}. Yet, those datasets were often small in scale (containing 200--1000 instances) and were not designed to test the reasoning process by which a metaphor is interpreted, which remains an open question. 

This paper presents a novel Metaphor Understanding  Challenge Dataset for LLMs (MUNCH). It provides over 10k paraphrases for metaphorical sentences and 1.5k triples of a metaphorical sentence and two candidate paraphrases, which could be apt or inapt (for dataset examples see Figure~\ref{fig:intro_dataset}; for statistics see Appendix~\ref{appx:prev_datasets}).
The metaphorical sentences were extracted from VUA texts, spanning four genres (academic, news, fiction, and conversation) and featuring metaphors at different levels of novelty.
Each metaphorical sentence contains a content word that is marked as metaphorically used.
A candidate paraphrase replaces the metaphorical word with another word, so that the resulting sentence is the same as the reference sentence except for that one word, therefore representing a lexical substitution task.
An apt paraphrase shows correct contextual interpretation of the metaphor while an inapt paraphrase uses a word that is related to 
a literal, source domain sense of the metaphorical word
(see the examples of \textcolor{Green}{correct} and \textcolor{red}{incorrect} substitution words in Figure~\ref{fig:intro_dataset}).
Such a setup of the task is inspired by the conceptual metaphor theory \cite{lakoff-johnson-1980-metaphors} and allows us to investigate whether the model performs full metaphor interpretation by cross-domain mapping or rather resorts to more shallow lexical similarity. In order to investigate this in a more controlled fashion, we opted for a lexical substitution task. Specifically, we test whether the model consistently chooses the correct target domain paraphrase (therefore, fully interpreting the metaphor) or rather bases its decisions on lexical similarity and chooses the inapt paraphrase that is similar in meaning to the literal use of the metaphorical word.

We set up a fill-in-the-blank task to crowdsource apt paraphrases, and manually selected the best paraphrases using expert knowledge.
We also manually created inapt paraphrases from WordNet synsets, so that the apt and inapt paraphrases reflect the target and source domains of the metaphors respectively. Specifically, the inapt paraphrases are synonyms or hypernyms associated with the word's literal use (the source domain).

Using this dataset, we tested the metaphor understanding capabilities of
LLaMA-13B, 30B, and GPT-3.5 zero-shot in two tasks: paraphrase judgement and paraphrase generation, as illustrated in Figure~\ref{fig:intro_tasks}.\footnote{The dataset and the code will be released on GitHub and will be freely accessible.}
Our results show that both tasks are challenging for the models.
In particular, the models are prone to confuse the target and source domains of the metaphors, as they often fail to distinguish the inapt paraphrases from the apt paraphrases or reference sentences.
Our experiments also reveal that LLMs' metaphor understanding capabilities are associated with genre, metaphor novelty, and POS of the metaphorical word.
The MUNCH tasks thus allow us to gain insight into how LLMs process metaphors as well as how this remarkable ability can be improved in the future.

%% file: body/2_related_work.tex
\citet{steen-etal-2010-mipvu} created VUA, which marks out \textbf{metaphor-related words (MRWs)} 
in a 4-million-word subset of the British National Corpus.
MRWs are lexical units implicative of underlying cross-domain mappings;
they can be directly or indirectly used, depending on their contextual meaning. 
Consider the sentence \enquote{A small five-year-old \underline{perched} like a \underline{mosquito} on the beginners' pony}.
The noun \lingform{mosquito}
is a \textbf{direct metaphor},
as it literally means mosquito (the source domain) in this context.
The verb \lingform{perched}
is an \textbf{indirect metaphor},
because it has a more basic usage, as in
\enquote{A pair of glasses were perched on the bridge of his nose}.

VUA has been widely used in studies on automated metaphor detection \cite{leong-etal-2018-report,leong-etal-2020-report,choi-etal-2021-melbert,zhang-liu-2022-metaphor,li-etal-2023-framebert}.
However, the corpus
does not specify the conceptual metaphors indicated by the MRWs or provide annotation for interpreting the metaphors.
The corpus is not directly applicable to automated metaphor interpretation.

\citet{shutova-2010-automatic} defined automated metaphor interpretation as a paraphrasing task:
Given a metaphorical expression where a word is marked as metaphorically used, the model should replace this word with another word to render a literal paraphrase of the expression.
For example, the verb phrase \lingform{stir excitement}, where \lingform{stir} is used metaphorically, should be paraphrased as \lingform{provoke excitement}.

\citet{bizzoni-lappin-2018-predicting} created a Metaphor Paraphrase Evaluation Corpus (MPEC), which provides correct and incorrect paraphrases for $\sim$200 short sentences containing metaphor use;
paraphrases could greatly differ from the reference sentences.
\citet{joseph-etal-2023-newsmet} created the NewsMet dataset, which consists of 1k verbal metaphors in news headlines as well as their literal equivalents; incorrect paraphrases are not provided.

Several recent studies approached metaphor understanding
as an inference task.
The IMPLI dataset \cite{stowe-etal-2022-impli} 
includes entailed and non-entailed sentences for $\sim$900 metaphorical sentences.
The FLUTE dataset \cite{chakrabarty-etal-2022-flute}
provides entailment and contradiction pairs for 1500 metaphorical sentences (including 750 similes).
Fine-tuned transformer-based models reached > 0.8 accuracies
in these 2 studies in predicting the class of a given sentence pair.

Recent studies also employed multiple-choice and generative tasks
to assess LLMs' ability to reason with metaphorical language.
The MiQA benchmark \cite{comsa-etal-2022-miqa}
uses such tasks to test whether
models can distinguish metaphorical and literal uses of the same words;
150 conventional metaphors are involved.
The Fig-QA task \cite{liu-etal-2022-testing}
includes 10k similes (a type of direct metaphor)
and requires models to distinguish a pair of metaphors of opposite meanings.
\citet{chakrabarty-etal-2022-rocket}
examined LLMs' figurative language understanding
by asking them to generate text after encountering an idiom or simile.

The MUNCH dataset
provides 3k samples of indirect metaphors, 10k correct paraphrases, and 1.5k incorrect paraphrases.
It is therefore one of the largest datasets for paraphrasing of indirect metaphors.
The candidate paraphrases are also systematically different from the ones in previous datasets, as we tailored the dataset for testing LLMs' understanding of metaphors as cross-domain mappings and correctly capturing the underlying relational structures.
We summarise differences between MUNCH and previous datasets and provide more details for the latter in 
Appendix~\ref{appx:prev_datasets}.

%% file: body/3_data_collection.tex
The metaphor samples in our dataset were selected from the publicly available metaphor corpus VUA.
Each metaphor sample is a sentence containing a highlighted MRW, the metaphorical word to be interpreted; a paraphrase uses a single word to substitute the metaphorical word.
We use two criteria for selecting metaphorical sentences: novelty of the metaphor and possibility of single-word substitution.    
We explicate our selection process below. 

\paragraph{The novelty criterion.}
We employed novelty scores
from \citet{do-dinh-etal-2018-weeding}
to increase the proportion of novel metaphors in our dataset. Scores range from -1 to 1.
VUA contains a large proportion of conventional metaphors: The metaphorical use of the word can be found in a dictionary of contemporary language use \citep{steen-etal-2010-usage}.
As LLMs might have encountered enough data for such conventional metaphor uses during pre-training, the understanding of such metaphors should be relatively easy.
To render a more challenging dataset, we excluded MRWs with novelty scores below -0.3.
Metaphors with a novelty score higher than -0.3 could still be conventional:
The crowd workers who provided the novelty annotations 
in \citet{do-dinh-etal-2018-weeding}
relied on their intuition instead of a dictionary like \citet{steen-etal-2010-usage}.
And metaphorical uses included in dictionaries may still be considered novel by lay people.
We chose -0.3 as the threshold in order to collect a large and diverse dataset as a starting point.

\paragraph{The single-word criterion.}
To ensure that the metaphorical sentences can be paraphrased via single-word substitution, we excluded MRWs that are marked as direct metaphors, as well as a portion of indirect metaphors.
Directly used MRWs usually occur in a sequence, such as ``I knew the pathway like the \underline{back of my hand}''.
They are thus not suitable for single-word substitution.
Also, the direct metaphor \lingform{back of my hand} refers literally to the back of the speaker's hand---its contextual meaning is directly associated with the source domain.
This is contrary to our task setup, where apt paraphrases (contextual meaning) should be associated with target domains.
We therefore opted to focus on indirect metaphors in this study.
    
Within the category of indirect metaphors, we filtered out new-formations, consecutive MRWs, and proper names.
New-formations are words that do not have an entry in dictionary, so VUA annotated the parts that do have corresponding entries.
For example, in the phrase \lingform{a rose-tinted vision of the world}, the word \lingform{rose-tinted} was a new-formation; so \lingform{rose} and \lingform{tinted} are marked as separate MRWs in VUA and received their separate novelty scores. 
We filtered these out because a single metaphorical word should have a single novelty score (\lingform{rose-tinted} has two), yet it is hard to paraphrase \lingform{rose} or \lingform{tinted} instead of \lingform{rose-tinted} altogether.

Likewise, we excluded cases where multiple content words marked as indirect metaphors occur consecutively, such as \lingform{take place}, \lingform{long road home}, \lingform{great leap forward}.
These often involve fixed expressions or phrases that either should be replaced as a whole or should not be marked as consecutive indirect metaphors.
We also excluded metaphorical words that are part of a proper name, which,
like fixed expressions, need to be treated as a whole.
For example, the proper name \lingform{Nord Stream} would lose its meaning if one changed the metaphorical word \lingform{stream} into another word.

%% file: body/4_ann_apt.tex
\paragraph{Crowdsourcing task.}
We constructed a fill-in-the-blank task to crowdsource (apt) paraphrases for the metaphorical sentences.
Each task included 30 sentences to be paraphrased, so that the task can be finished within 30 minutes.
Under each sentence, the workers were presented with a copy of the sentence where the metaphorical word is replaced with a blank; they were asked to fill the blank with a single word so that the new sentence is a semantically and grammatically apt paraphrase of the reference sentence.
If they were not able to paraphrase the sentence, they were asked to explain why it was difficult.
Examples of good and bad answers were provided in the instructions (see Appendix~\ref{appx:pptask}).

The workers were recruited via {Prolific}\footnote{https://www.prolific.co/}.
We set prescreening criteria to only include adult (age > 18) native English speakers who were living in an English-speaking country and did not have any language-related disorders.
The workers were asked to confirm within the task that they met these criteria.
After giving consent to participate and reading the instructions, they were also required to correctly paraphrase a trial sentence in order to access the task.
More details (worker's consent, the trial sentence) are given in Appendix~\ref{appx:pptask}.

We released 99 tasks in total and collected
5 data points for each of the 2970 reference sentences. 
We received single-word substitutions for 2953 sentences (the other 17 are presented and explained in Appendix~\ref{appx:pptask}), and 61\% of them got repeated answers---multiple workers submit the same paraphrase despite the question being open-ended.
This confirms the reliability of our task.

\paragraph{Expert validation.}
For a selection of the reference sentences for which we later annotated inapt paraphrases (Section~\ref{sec:ann_inapt}), we further validated the crowdsourced paraphrases to determine the best paraphrase for creating the triples (one metaphor sample, two candidate paraphrases).

We used
both majority vote and expert knowledge to find one best paraphrase for each reference sentence.
For each sentence, we first sorted the collected single-word substitutions from the most to the least popular (in terms of how many participants proposed that substitution).
The apt paraphrase that was proposed by the highest number of participants was verified by the authors and selected as the best paraphrase for that reference sentence.

When multiple apt paraphrases have the same number of votes, we chose the one that is clearly within the target domain---that is, the paraphrase clearly shows that the metaphorical word is interpreted in its contextual sense.
For instance, 
we received 5 different single-word substitutions for the metaphorical word \lingform{attack}
in the sentence ``\ldots he has become involved in a row over his \underline{attack} on the \enquote{Pharisees} of British society''.
These are
\lingform{remarks}, \lingform{views}, \lingform{offense}, \lingform{incursion}, and \lingform{disagreement}, each proposed by a single participant.
All of them can be considered apt paraphrases.
We chose \lingform{remarks} because it clearly shows the metaphorical word \lingform{attack} is interpreted in the \concept{argument} domain.
The meaning of \lingform{offense} and \lingform{disagreement} are more abstract and could involve other conceptual domains;
the paraphrases that replace \lingform{attack} with \lingform{views} and \lingform{incursion} respectively are still metaphorical, as \lingform{view} can be associated with \concept{vision} and \lingform{incursion}, like the metaphorical word itself, is still in the domain of \concept{battle}.
These four are thus less preferable with regard to the purpose of our dataset.

While we managed to find one best paraphrase for most reference sentences, there are 45 for which we selected two paraphrases as the best, as the two received the same votes and are equally apt.
There are also 11 sentences for which no paraphrase was selected.
These are cases where the given context is insufficient for determining the contextual meaning of the metaphorical word.

%% file: body/5_ann_inapt.tex
\citet{tong-2021-metaphor} shows that incorrect paraphrases based on the basic sense of the metaphorical word are the least distinguishable from correct ones (i.e., paraphrases based on the contextual sense) with respect to aptness.
To render truly challenging inapt paraphrases for our task, we therefore created inapt paraphrases exclusively from basic senses.

We employed WordNet for identifying basic senses and obtaining sense-specific synonyms,
following the annotation guidelines presented in Appendix~\ref{appx:inapt}.
For each metaphorical word, we first locate the WordNet synsets that correspond to its more basic meaning (relative to its contextual meaning in the reference sentence).
Then we go through the synonyms (or hypernyms when no synonyms are provided) under the basic-sense synsets and select those that are clearly associated with the metaphor's source domain and would render a grammatical (but inapt) paraphrase.

We went through all 2970 sentences released for the crowdsourcing task and found inapt paraphrases for 991 of them.
After removing items lacking apt paraphrases (either because no single-word substitutions were crowdsourced or because none of the collected ones are of sufficient quality), we created 1492 triples for 728 metaphorical sentences, including 1072 triples with an apt and an inapt paraphrase, 375 triples with two inapt paraphrases, and 45 with two apt paraphrases.

\paragraph{Inter-annotator agreement.}
From the 991 sentences for which inapt paraphrases were identified,
we randomly selected 200 to be annotated by a second annotator.
The annotator was a PhD candidate in linguistics specialising in metaphor research.
We explained the annotation process to the second annotator through a meeting and the guidelines  in Appendix~\ref{appx:inapt}.
The Gwet's gamma coefficient for the agreement between the two expert annotators is 0.84.

%% file: body/6_data_analysis.tex
\begin{table}[!t]
    \centering
    \small
    \begin{tabular}{c|c|c|c|c|c}
    & \texttt{ACPROSE} & \texttt{NEWS} & \texttt{FICTION} & \texttt{CONVRSN} & TOTAL \\
             \hline

    & 1061 &  922 &  593 &  377 & 2953 \\
         \hline

    N     & 50\% & 39\% & 35\% & 25\% & 40\% \\
    V     & 35\% & 42\% & 39\% & 51\% & 40\% \\
    A     & 16\% & 20\% & 26\% & 24\% & 20\% \\
    \end{tabular}
    \caption{Number of metaphor samples per genre (academic, news, fiction, conversation), and the percentage of sentences where the metaphorical word is a noun (N), a verb (V), or either an adjective or an adverb (A).}
    \label{tbl:dataset-counts}
\end{table}

The dataset contains approximately the same number of samples from academic and news genres, and fewer samples from fiction and conversation, as shown in Table \ref{tbl:dataset-counts}.
These metaphor samples cover metaphorical use of content words in all four parts of speech.
Noun and verb MRWs are of a higher proportion compared to adjectives and adverbs.
In news and fiction, these two categories have similar percentages.
The academic genre contains more noun MRWs than verbs whereas in conversation the situation is reversed: Half of the metaphorical words are verbs, while the percentage of nouns is similar to that of adjectives and adverbs.

As we excluded MRWs of novelty scores lower than -0.3, the metaphor samples exhibit a wider range of novelty scores above 0 than below 0 (see Appendix~\ref{appx:novelty}).
Meanwhile, a large proportion of the metaphor samples could be considered only slightly novel or conventional (novelty scores between -0.3 and 0.3).
Metaphor samples of the highest novelty scores can be from any of the four genres.
Despite their different proportion in the entire dataset (Table~\ref{tbl:dataset-counts}), all four genres include metaphor samples across all levels of perceived novelty.

\begin{figure}[!t]
    \centering
    \includegraphics[width=0.9\linewidth]{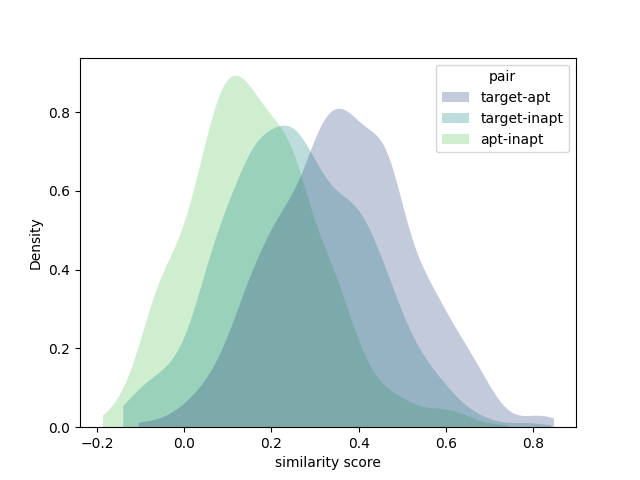}
    \caption{Distribution of the cosine similarity between target-apt, target-inapt, and apt-inapt pairs.}
    \label{fig:triples-cossim}
\end{figure}

We also examined the cosine similarity between the metaphorical words and apt and inapt substitutions.
Since the inapt paraphrases were based on the more basic meaning of the MRWs (section~\ref{sec:ann_inapt}), we expected inapt substitution words to be more similar to the metaphorical words than apt substitution words.
We computed the cosine similarity scores using \texttt{glove-wiki-gigaword-300} embeddings \cite{pennington-etal-2014-glove}, accessed through the \texttt{gensim} Python library.
Figure~\ref{fig:triples-cossim} shows the distribution of cosine similarity scores for 1006 triples, excluding the ones containing out-of-vocabulary words.
Surprisingly, the target-apt pairs tend to have higher cosine similarity scores than the target-inapt pairs.
The plot suggests that the 3 pairs are quite distinguishable in terms of cosine similarity scores, with target-apt pairs being the most similar, and apt-inapt the least similar.
This might be associated with the fact that our metaphorical sentences were sampled from VUA, which, being representative of metaphor use in natural discourse, includes a large percentage of conventional metaphors.
Nonetheless, the majority of the cosine similarity scores are above 0, and the 3 pairs still share a wide range of similarity scores.
The distribution plot is therefore also suggestive of the reliability of our dataset, as well as its potential challenge for LLMs.

%% file: body/7_model_eval.tex
We evaluated LLaMA-13B, LLaMA-30B, and GPT-3.5 (\texttt{text-davinci-003}) on two tasks: (Task~1)
\textbf{paraphrase judgement}, which requires a model to select correct paraphrases for a given reference sentence from given candidates; and (Task~2)
\textbf{paraphrase generation}, which asks a model to generate correct paraphrases for a given reference sentence. 
The paraphrase judgement task used the 1492 triples that include inapt paraphrases; 
the generation task used all 2953 metaphorical sentences.
Details regarding computational budget is given in Appendix~\ref{appx:prompts}.

\subsection{Paraphrase judgement}

\begin{figure}[!t]
    \centering
    \includegraphics[scale=0.7]{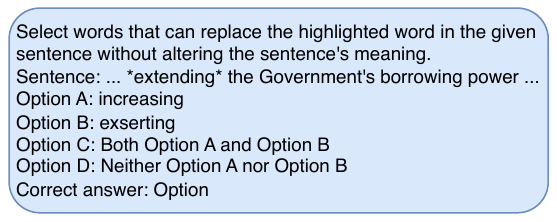}
    \caption{\label{fig:ppjudge}
    Example prompt for the \textit{Word-judgement} task (the \textit{Implicit} condition).
    The given sentence is shortened for illustration.}
\end{figure}

We evaluate the LLMs in a prompting setup. We test the models' ability to interpret metaphor under different conditions. In the first scenario, we prompt the model by providing the reference sentence with the metaphorical word highlighted and two candidate replacement words for it (\textit{Word-judgement}). In the second scenario, each of the candidate replacement words is embedded in the sentence (\textit{Sentence-judgement}). In both cases the model needs to solve a multiple choice task. Besides providing the apt and inapt paraphrases (Options A and B) as answer options, we also complement them with Option C, that both candidates are correct, or Option D, that neither are correct. See Figure~\ref{fig:intro_tasks} for an example. We expect \textit{Word-judgement} to be more challenging, as the model would need an additional inference step compared to sentence judgement, to replace the metaphorical word with the two given options and (implicitly) form the intended paraphrases. 

For both \textit{Word-judgement} and \textit{Sentence-judgement} setups, we further investigate whether it makes a difference if the model is explicitly ``told'' that the task is to paraphrase a metaphor or not. This results in three further conditions: \textit{Implicit} (not mentioning metaphor in the prompt), \textit{Metaphor-Sent} (revealing that the reference sentence contains a metaphor), and
\textit{Metaphor-Word} (revealing that the specific highlighted word in the sentence is metaphorically used). The \textit{Implicit} condition corresponds best to the real-life application of LLMs, where the model needs to be able to interpret metaphors without being instructed that metaphors are there.

We tested LLaMA-13B and 30B, and GPT-3.5 in each of the 6 conditions, using 3 prompts for each condition (the prompts are listed in Appendix~\ref{appx:prompts}).
Table~\ref{tbl:4opt-results} shows the mean accuracy and standard deviation for each model in each condition.
The random baseline achieves an accuracy of 0.25, as there is always only one correct option out of the given four.
\begin{table}[!t]
\centering
\small
\begin{tabular}{lccc}
\hline
           & LLaMA-13B & LLaMA-30B & GPT-3.5   \\
\hline
Word-judge &           &           &           \\
Implicit   & .28 (.18) & .21 (.10) & .23 (.10) \\
M-Sent     & .30 (.16) & .19 (.09) & .20 (.10) \\
M-Word     & .33 (.18) & .21 (.08) & .20 (.08) \\
\hline
Sent-judge &           &           &           \\
Implicit   & .13 (.06) & .14 (.03) & .17 (.07) \\
M-Sent     & .12 (.07) & .17 (.03) & .16 (.06) \\
M-Word     & .10 (.08) & .27 (.05) & .21 (.02) \\
\hline
\end{tabular}
\caption{\label{tbl:4opt-results}
Mean (SD) accuracies across 3 prompts for each paraphrase judgement condition.}
\end{table}
The performance of all three models was below the random baseline in most cases, except for LLaMA-13B in the \textit{Word-judgement} tasks and LLaMA-30B in the \textit{Metaphor-Word} condition of the \textit{Sentence-judgement} task.
Meanwhile, the accuracy of LLaMA-13B varied a great deal across different prompts in the \textit{Word-judgement} tasks. 

The \textit{Sentence-judgement} task seems to be more challenging than \textit{Word-judgement} for the models.
For LLaMA-30B and GPT-3.5, the task was particularly difficult when they were not instructed to focus on the metaphorical word, and were not informed that the word is metaphorically used (the \textit{Metaphor-Word} condition).
For LLaMA-13B, all 3 \textit{Sentence-judgement} conditions are similarly difficult.
However, its higher accuracies in the \textit{Word-judgement} tasks also indicate the benefit of instructing the model to focus on the metaphorical word.

\subsection{Paraphrase generation}

\begin{figure}[!t]
    \centering
    \includegraphics[scale=0.7]{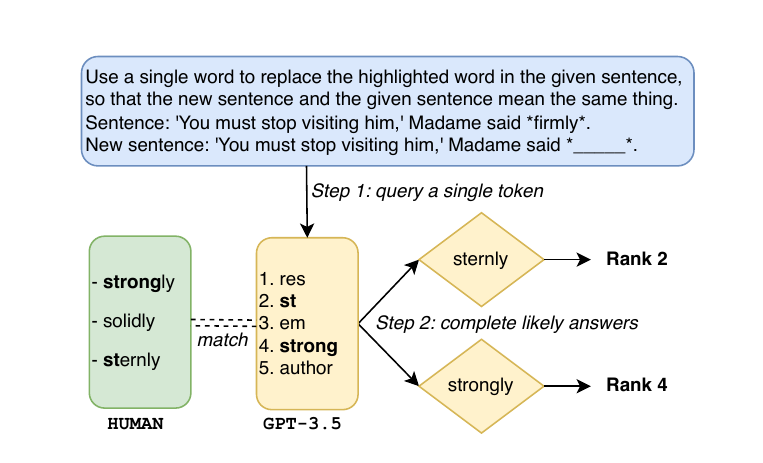}
    \caption{\label{fig:ppgen-flowchart}
    Procedure of the paraphrase generation task, using GPT-3.5 prompt and outputs as example.
    We first ask the model to generate a single token to get a glimpse of its top 5 answers.
    For each token that matches the beginning of a human answer, we let the model complete it to see whether it is a complete match.}
\end{figure}

The purpose of this task is to compare model and human performance in paraphrasing metaphorically used words.
The prompts were thus designed to be semantically close to the instructions in our crowdsourcing task (Section~\ref{sec:ann_apt}). The model answers were generated in two steps (see Figure~\ref{fig:ppgen-flowchart}).
We first let the models generate a single token---this allowed us to access the models' ranking of all tokens in their vocabulary. Of these, we selected the ones that match human annotations and let the models complete them into words.
The completions were then compared with human annotations to determine the rank of each expected answer.

\begin{table}[!t]
\centering
\small
\begin{tabular}{lccc}
\hline
          & MRR       & Recall@5  & Recall@10    \\
\hline
LLaMA-13B & .33 (.02) & .22 (.02) & .33 (.02)    \\
LLaMA-30B & .47 (.03) & .28 (.02) & .40 (.03)    \\
GPT-3.5   & .54 (.02) & .32 (.01) & -            \\
\hline
\end{tabular}
\caption{\label{tbl:ppgen-results}
Mean (SD) performance across 3 prompts in the paraphrase generation task.
Recall@10 does not apply to GPT-3.5 as the OpenAI API only allows access to the top-5 answers.}
\end{table}

We tested the models on 3 prompts (see Appendix~\ref{appx:prompts}) and their mean performance in terms of mean reciprocal rank (MRR), recall at top 5 paraphrases and recall at top 10 paraphrases is shown in Table~\ref{tbl:ppgen-results}.
GPT-3.5 performed best and LLaMA-30B came second.
The models' performance was also more stable across different prompts as compared to the paraphrase judgement task.
Nonetheless, all three models clearly preferred different answers as compared to human annotators.

%% file: body/8_discussion.tex
\paragraph{Paraphrase judgement.}
We looked into the type of errors the models made in paraphrase judgement.
The number of each combination of expected and predicted answers for each model is in Appendix~\ref{appx:error}.
We found that LLaMA-30B and GPT-3.5 could ignore the semantic differences between a given sentence and an inapt paraphrase, as they tend to predict both candidates as correct when presented with one or more inapt paraphrases.
LLaMA-13B, on the other hand, tends to assume that the two given candidates always contain one apt and one inapt paraphrase.
Nonetheless, it did not seem capable of distinguishing the two, as it made a similar number of Option A and Option B predictions.

\paragraph{Paraphrase generation.}
We examined the top-ranked answers of the models and found 4 categories that the `incorrect' or unexpected answers could fall into.
\textbf{1) Nonsensical:} For the sentence ``...for this number \underline{line} I would say...'', GPT-3.5 gives \lingform{thus} as the best substitution word, ignoring the meaning of \lingform{line}, whereas LLaMA-13B repeats \lingform{number}. \textbf{2) Lack of contextual understanding:} In ``...he touched both \underline{sides} of the coin...'', GPT-3.5 replaces the word \lingform{sides} with \lingform{facets}, suggesting that it neglects details of the meaning of the sentence (that a coin only has 2 sides).
\textbf{3) Ungrammatical:}
On the other hand, the model may have understood the metaphor, but fails to convert its understanding into a suitable substitution word. In ``...they all \underline{shared} the emphasis on `her'...'', LLaMA-30B suggests \lingform{concurred} as the best answer, implying that the meaning of \lingform{shared} is understood, but that grammatical agreement has been sacrificed.
\textbf{4) Preference:} Finally, the disagreement between the models and human annotators may simply be a matter of preference. For ``For a man whom Rebecca West ... called `\underline{repulsive}' and `treacherous'...'', crowd workers provide 4 possible answers: \lingform{revolting}, \lingform{disgusting}, \lingform{awful}, and \lingform{grotesque}.
Both LLaMA-30B and GPT-3.5 give \lingform{odious} as the best answer.
Here, both the human annotators and the models understand and can paraphrase the sentence, and it is hard to say whose answer is best.

\paragraph{Factors associated with model performance.}
We also examined the association between model performance and 3 factors: genre, metaphor novelty, and the POS of the metaphorical word.
The details are available in Appendix~\ref{appx:error}.
We found metaphors of higher novelty scores to be more difficult for LLaMA-30B in paraphrase judgement, and for GPT-3.5 in paraphrase generation.
The association between genre or POS and model performance tends to differ per model and task.
The fiction genre, for example, is the easiest for the LLaMA models in paraphrase generation; yet it is the most difficult for GPT-3.5 in the generation task and for LLaMA-30B in the judgement task.
Similarly, noun metaphors are the easiest for LLaMA-30B in the generation task and for GPT-3.5 in the judgement task.
Meanwhile for GPT-3.5 in the generation task, adverb metaphors become the easiest.

To sum up,
the results of the two paraphrase tasks indicate that the LLMs are unable to (fully) understand some of the metaphors in our dataset.
The paraphrase judgement task further reveals that the models have difficulty distinguishing the metaphors' source domains (implied by the inapt paraphrases) and target domains (implied by the reference sentences and apt paraphrases). This further suggests that the models are unlikely to perform reasoning across semantic domains; when they succeed in understanding the metaphor, they may still reason in ways that are different from humans.
This means that for downstream NLP tasks such as opinion mining, bias detection, humour detection, and intent recognition,
the LLMs could overlook the entailment of a metaphor.
In machine translation as well as summarisation of highly figurative or poetic texts,
the problems may manifest as incorrect or peculiar explanation of metaphors.

A direction for improvement is to mark out metaphor uses in texts and direct the model's attention to them:
In the paraphrase judgement task, the models reach higher accuracies when the metaphorical word is marked out (in the \textit{Word-judgement} task or in a \textit{Metaphor-Word} condition).
However, since the models generally performed poorly in the experiments, the LLMs may need to be fine-tuned in order to better understand metaphors.
When fine-tuning, one can consider increasing the proportion of certain metaphor types in training data, as genre, metaphor novelty, and POS of the metaphorical word are all associated with model performance.
Future studies could first employ MUNCH to detect the weak points of an LLM and then curate training data accordingly.

%% file: body/a2_related_datasets.tex
\begin{table*}[!t]
\centering
\begin{tabular}{lrcrcrc}
\hline
      & \#met & met:length & \#correct & correct:type      & \#distractor & distractor:type \\
\hline
MPEC  &   192 &  9  (4)    &       218 & $s \rightarrow s$ &          526 & mixed \\
NewsMet & 791 & 12  (3)    &       791 & $w \rightarrow w$ &            0 & NA \\
IMPLI &   913 & 16 (10)    &      1032 & $w \rightarrow p$ &          281 & context change \\
FLUTE &  1500 & 11  (5)    &      1500 & $p \rightarrow p$ &         1500 & opposite meaning \\
MiQA  &   150 &  8  (2)    &       150 & $s \rightarrow s$ &          150 & context change \\
Fig-QA & 10256 & 9  (3)    &     10256 & $s \rightarrow s$ &        10256 & opposite meaning \\
\textbf{MUNCH}
      &  2953 & 26 (15)    &     10261 & $w \rightarrow w$ &         1492 & paraphrase \\
\hline
\end{tabular}
\caption{\label{tbl:munch_vs_prev}
Differences between MUNCH and previous datasets that provide paraphrases for metaphors: MPEC (\citealp{bizzoni-lappin-2018-predicting}; \url{github.com/yuri-bizzoni/Metaphor-Paraphrase}), NewsMet (\citealp{joseph-etal-2023-newsmet}; \url{https://github.com/AxleBlaze3/NewsMet_Metaphor_Dataset/tree/main}), IMPLI (\citealp{stowe-etal-2022-impli}; \url{github.com/UKPLab/acl2022-impli}), FLUTE (\citealp{chakrabarty-etal-2022-flute}; \url{https://github.com/tuhinjubcse/model-in-the-loop-fig-lang}), MiQA (\citealp{comsa-etal-2022-miqa}), and Fig-QA(\citealp{liu-etal-2022-testing}; \url{https://github.com/nightingal3/Fig-QA/tree/master}). We present their differences regarding number of metaphor samples (\#met), mean (SD) length of the metaphor samples (met:length, measured by number of words), number of correct paraphrases (\#correct), the part of a metaphor sample that is replaced to create correct paraphrases (correct:type; $s$=sentence, $p$=phrase, $w$=word), number of distractors (\#distractor), and distractor type. The numbers are calculated from the datasets available on GitHub.
Note that our dataset is much more extensive than the previous ones.}
\end{table*}

Table~\ref{tbl:munch_vs_prev} summarises the differences between MUNCH and previous datasets.

Example~\ref{eg:mpec} is extracted from MPEC.
The correct paraphrase, sentence~\ref{eg:mpec-apt}, is almost completely different from the original sentence.
The two distractor sentences that follow indicate different types of misinterpretation:
Sentence~\ref{eg:mpec-inapt} wrongly interprets the meaning of the original sentence,
while the last sentence is based on a literal use of the word \lingform{wheels}.

\ex. \label{eg:mpec} the wheels of justice turn slowly
    \a. \label{eg:mpec-apt} it might take time but eventually justice prevails
    \b. \label{eg:mpec-inapt} {?`} justice prevails in very little time
    \b. {?`} the wheels of a car turn slowly
    
The MPEC corpus is employed by two metaphor understanding tasks in BIG-Bench \citep{srivastava-etal-2022-imitation}.
The \href{https://github.com/google/BIG-bench/tree/main/bigbench/benchmark_tasks/metaphor_boolean}{metaphor-boolean task} uses a binary classification setup:
Given a pair of sentences, is the second sentence a paraphrase of the first?
GPT-2 only reached 0.41 accuracy on this task in a zero-shot scenario.
The \href{https://github.com/google/BIG-bench/tree/main/bigbench/benchmark_tasks/metaphor_understanding}{metaphor-understanding task} 
consists of two subtasks: metaphor to paraphrase, which asks the model to select the correct paraphrase from 4 candidates;
and paraphrase to metaphor, which requires the model to distinguish the metaphorical sentence corresponding to a given paraphrase from 3 other metaphors.
GPT-2 large performed poorly on both subtasks: In a zero-shot scenario, the model gave 0.27 accuracy on the metaphor-to-paraphrase task, and 0.67 accuracy on the paraphrase-to-metaphor task.

The metaphor-literal pairs in the NewsMet dataset was created with the help of LLMs.
Each news headline has a verb considered as the focus word.
They first passed the headlines with the focus words masked to ALBERT \cite{lan-etal-2020-albert}
to obtain the first 200 words that can replace the focus word.
These 200 words were then passed to a metaphor detector to obtain the top-6 metaphorical and top-6 literal candidates.
Human annotators then identified the best
literal counterparts for metaphorical focus words and
the best metaphorical counterpart for literal focus words.

In the IMPLI example~\ref{eg:impli}, the correct paraphrase \ref{eg:impli-apt} uses a phrase, \lingform{paid for}, to explain the metaphorically used word \lingform{absorbed} in the original sentence.
The distractor, on the other hand, is based on the literal meaning of \lingform{absorbed}.
Fine-tuned RoBERTa base and RoBERTa large achieved high accuracies ($> 0.8$) on labelling these metaphor-paraphrase and metaphor-distractor sentence pairs.

\ex. \label{eg:impli} he absorbed the costs for the accident
    \a. \label{eg:impli-apt} he paid for the costs for the accident
    \b. {?`} he absorbed the sunlight after the accident

Example~\ref{eg:flute-met} is extracted from the FLUTE dataset;
included in the parentheses are explanations for the paraphrase and the contradict respectively.
Contrary to the MUNCH dataset, the authors aimed at paraphrases that use more than one word to replace a metaphorically used word.
Note that sentence~\ref{eg:flute-met-contradict} is more of a direct contradiction of the reference metaphor than the paraphrase, as it preserves the metaphorically used word \lingform{louder}.
The difference between the contradict and the reference metaphor may thus be easier to detect as compared to a contradict that is more similar to the paraphrase (e.g., \lingform{Actions are not more important than words}).

\ex. \label{eg:flute-met} Actions speak louder than words.
    \a. Actions are more important than words. (This phrase is used to say that what someone does is more important than what they say.)
    \b. {?`} \label{eg:flute-met-contradict} Actions are not louder than words. (The metaphor suggests that deeds or actions are more important than words, while the contradiction suggests that words are more important than deeds or actions.)

As example~\ref{eg:miqa} shows,
each metaphorical premise in the MiQA dataset is paired with a literal premise exemplifying literal use of the metaphorical word;
the dataset also includes implications (the text in parenthesis) of the metaphorical and literal premises:

\ex. \label{eg:miqa}
    \a. I \textbf{see} what you mean (I understand you)
    \b. I \textbf{see} what you are pointing at (My eyes are working well)

\citet{comsa-etal-2022-miqa} set up 2 binary-choice tasks using the MiQA dataset:
(1) Given a metaphorical premise, select the correct implication;
(2) given an implication, select the corresponding premise.
They also set up a generative task:
Given a metaphorical premise, answer whether it implies the literal conclusion.
LLMs performed well on these tasks.

The Fig-QA dataset provides similes of opposite meanings as well as their implications (given in parentheses):

\ex.
    \a. The meteor was as bright as New York City (The meteor was very bright)
    \b. The meteor was as bright as coal (The meteor was not bright at all)

The binary-choice task is similar to MiQA:
Given a metaphorical premise, select the correct implication.
They also develop a generative task which prompt models to generate implications freely.
\citet{liu-etal-2022-testing} found these tasks challenging
for LLMs in zero-shot settings.

%% file: body/a4_pptask.tex
The participant information sheet, which was presented to the crowd workers prior to the consent form, has a section dedicated to potential disadvantages and risks involved in participating in the study---

\blockquote{The sentences you will paraphrase were from a wide range of sources, including newspapers, fiction, and dialogues. You may occasionally encounter violence or taboo topics (e.g., war, crime, sex), as well as potentially disturbing opinions.

If you are concerned, you do not have to give consent; you can also withdraw anytime during the experiment.}

The information sheet also explains how data collected from the study will be used.
The workers were informed that their participation would remain confidential, that their response would be anonymised, and that the data would be made open access at the end of the study.

\begin{figure*}[!t]
    \centering
    \fbox{
    \includegraphics[width=15.5cm]{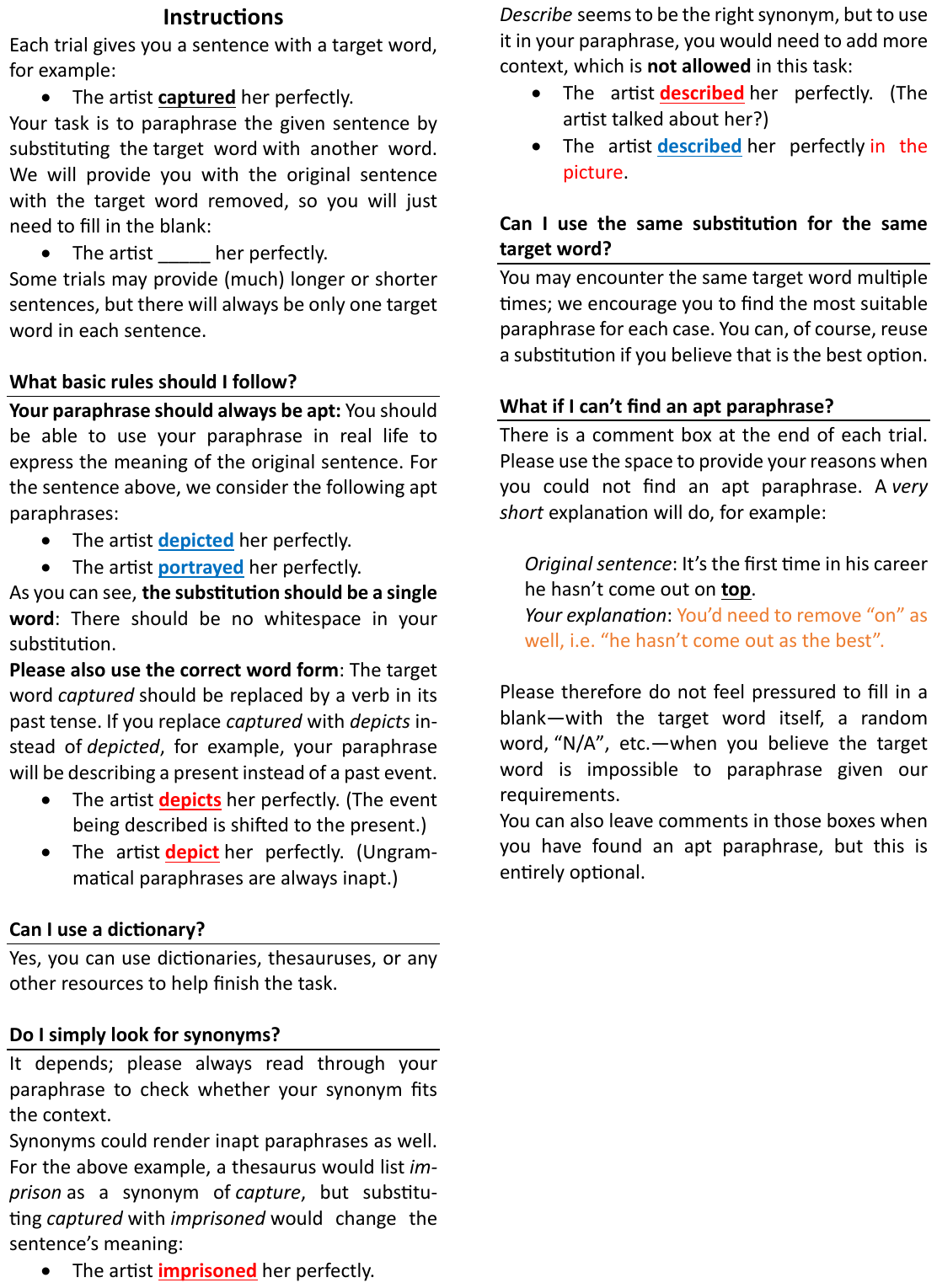}
    }
    \caption{Instructions for the paraphrasing task.}
    \label{fig:pptask-instruct}
\end{figure*}

The annotation guidelines are shown in Figure~\ref{fig:pptask-instruct}.
The trial sentence is provided in example~\Next, where \lingform{introduce} is the metaphorical word to be interpreted.
Our final list of acceptable answers includes:
\lingform{address}, \lingform{advance}, \lingform{clarify}, \lingform{convey}, \lingform{cover}, \lingform{define}, \lingform{describe}, \lingform{discuss}, \lingform{elucidate}, \lingform{establish}, \lingform{explain}, \lingform{mention}, \lingform{present}, \lingform{propose}, \lingform{reveal}, \lingform{share}, \lingform{show}, \lingform{state}, \lingform{submit}, \lingform{suggest}, \lingform{teach}, \lingform{unveil}.

\ex. I shall now \met{introduce} the concept of an elementary charge, 1.6 × 10 -19 C, carried by an elementary particle called the electron.

Table~\ref{tbl:nopp-sentences} presents the 17 sentences for which none of the crowd workers were able to provide single-word substitutions for the metaphorical words.
These are mainly highly conventionalised metaphors, for which it is usually difficult to find an alternative expression.
There are also cases where the target word is part of a multi-part word (e.g., \lingform{\underline{carry} out}, \lingform{\underline{point} of view}) or a phrase (e.g., \lingform{\underline{put} in an appearance}, \lingform{\underline{get} rid of}).
These stem from annotation mistakes in VUA: According to the MIPVU procedure,
VUA should have marked the entire word or phrase as a single annotation unit.
We still collected paraphrases for these cases as there were no suitable way to filter them out automatically.

\begin{table*}
\centering
\begin{tabular}{r p{14.5cm}}
\hline
1 & The summer's sprawl begins to be oppressive at this stage in the year and trigger fingers are itching to snip back overgrown mallows, clear out the mildewing foliage of \met{golden} rod and reduce the overpowering bulk of bullyboy ground cover. \\
2 & The red and green of the Aztec necklace links it compositionally with the indigenous plants to the ‘south’ of the painting, the pink colonial-style dress tonally blending with the skyscrapers to the ‘\met{north}’. \\
3 & Nine out of 10 are routine calls, many of which could be \met{carried} out by mini cabs. \\
4 & This example assumes that a sympathy for motorists with \met{overwhelm} any tendency to logical analysis. \\
5 & There were, in fact, about a \met{score}. \\
6 & Mrs Bottomley is convinced the Tory victory provides the opportunity to entrench the reforms — and to give doctors, nurses and managers the confidence to \met{make} them work. \\
7 & Thus, as with biological theories, crime is seen as pathological (a disease), as something to be looked at from the medical \met{point} of view. \\
8 & ‘So you've decided to \met{put} in an appearance?’ \\
9 & He was in \met{there} twice, at a Wimpole Street number and again at an address in Mill Hill: Rufus H. Fletcher, MB, MRCP. \\
10 & Once again he backtracks and assumes a larger unity in which conflict \met{takes} place. \\
11 & no I'm alright Ann, I mean, feel a bit ba ah I mean I'm sorry I do have to buy a \met{feel} a bit of, I feel a bit dizzy you know as if I \\
12 & Mick said to me last night, he said to me you can never \met{fit} not used to it, but \\
13 & Now if he doesn't get the economy right he's gon na end up with \met{egg} on his face and \\
14 & That \met{take} me nearly all the er \\
15 & As this is been shared by \met{lines} int it? \\
16 & Well seven nines, well ee er, it \met{takes} you so long \\
17 & Take what you want and leave the rest, your mother'll \met{get} rid of it. \\
\hline
\end{tabular}
\caption{\label{tbl:nopp-sentences}
Sentences that did not receive single-word substitutions in the crowdsourcing task.}
\end{table*}

%% file: body/a5_inapt.tex
\begin{figure*}[!t]
    \centering
    \rotatebox{90}{
    \fbox{
    \includegraphics[width=23cm]{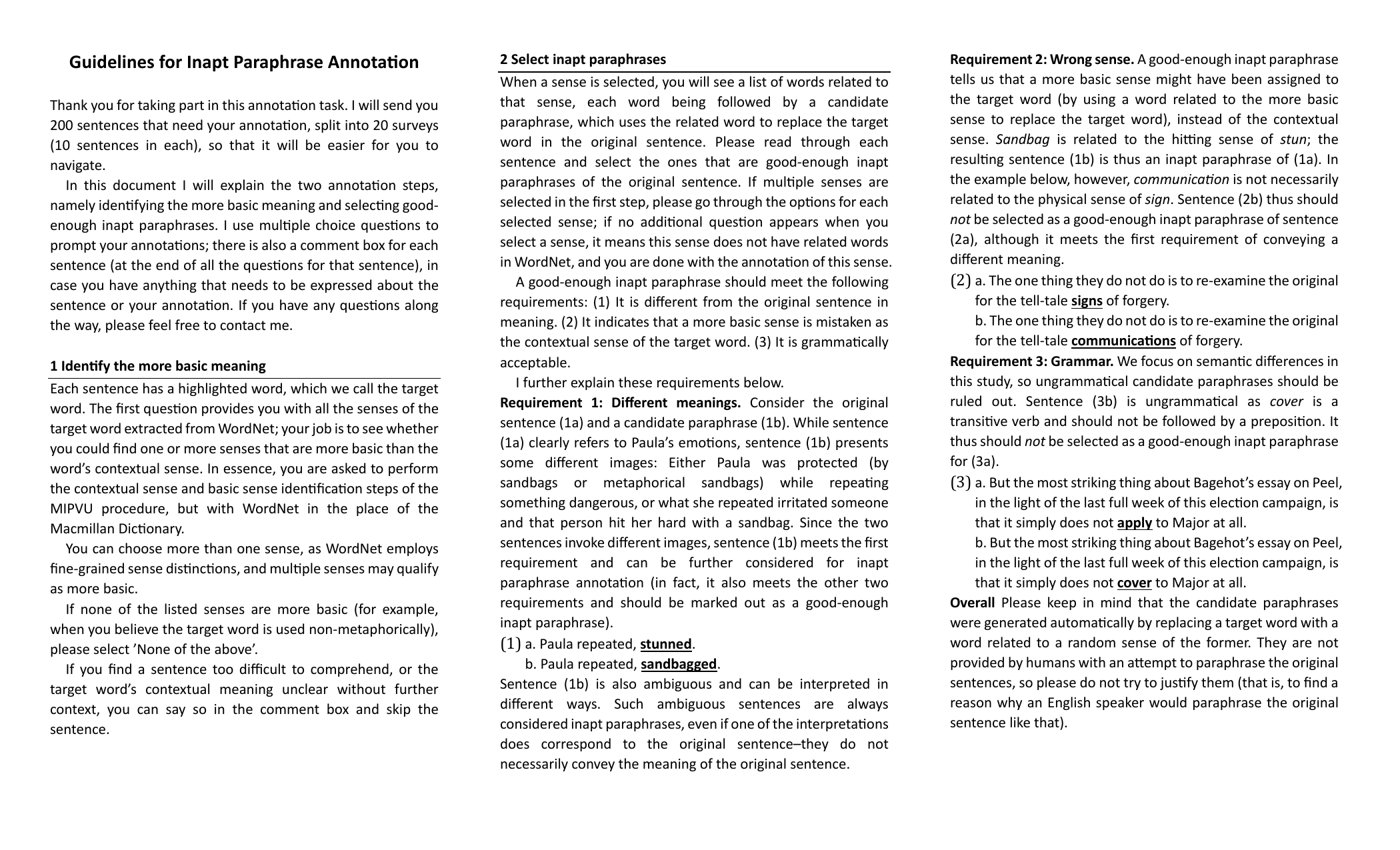}
    }
    }
    \caption{Guidelines for inapt paraphrase annotation.}
    \label{fig:inapt-guidelines}
\end{figure*}

The guidelines for inapt paraphrase annotation are presented in Figure~\ref{fig:inapt-guidelines}.

%% file: body/a6_novelty.tex
\begin{figure}[!t]
    \centering
    \includegraphics[width=0.9\linewidth]{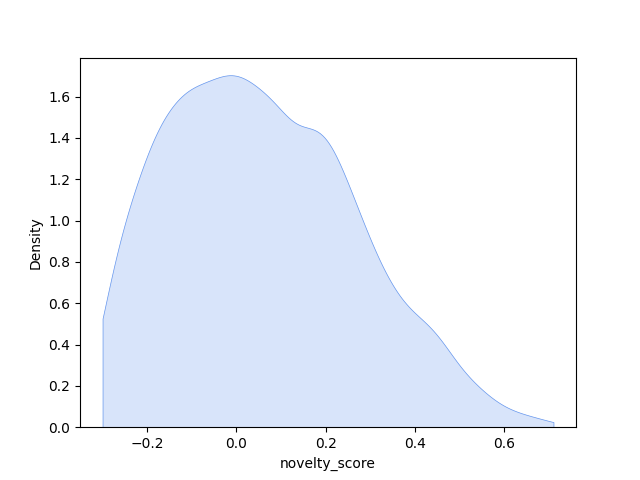}
    \includegraphics[width=0.9\linewidth]{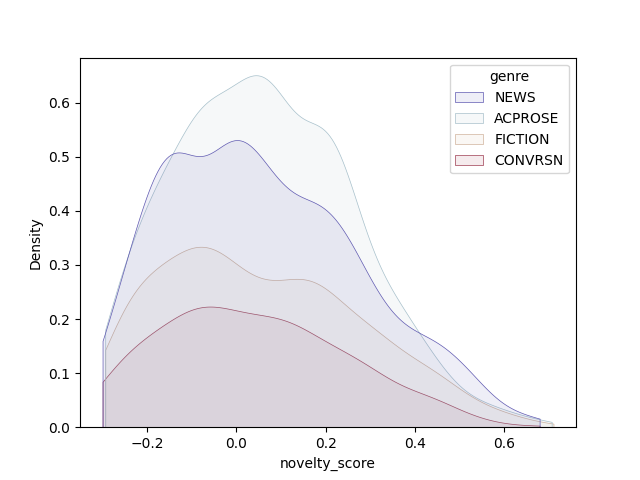}
    \caption{Novelty distribution of the metaphor samples, across all genres (above) and in different genres (below). The novelty scores are extracted from \citet{do-dinh-etal-2018-weeding}.}
    \label{fig:novelty}
\end{figure}

Figure~\ref{fig:novelty} presents the novelty distribution of the metaphor samples in MUNCH.

%% file: body/a7_prompts.tex
\begin{table}[!t]
\small
\centering
\begin{tabular}{p{0.45\textwidth}}
\hline
\textbf{Implicit} \\
\hline
(1) Choose the word(s) that can replace the highlighted word in the given sentence without changing the meaning of the sentence. \\
Sentence: (a metaphor sample) \\
Option A: (a substitution word) \\
Option B: (another substitution word) \\
Option C: Both Option A and Option B \\
Option D: Neither Option A nor Option B \\
Correct answer: Option \\
\hline
(2) Select words that can replace the highlighted word in the given sentence without altering the sentence's meaning.
(\ldots) \\
\hline
(3) Which of the given options can replace the highlighted word in the given sentence without altering the sentence's meaning?
(\ldots) \\
\hline
\textbf{M-sent} \\
\hline
(1) Choose the word(s) that can replace the highlighted word in the given metaphorical sentence without changing the meaning of the sentence.
(\ldots) \\
\hline
(2) Select words that can replace the highlighted word in the given metaphorical sentence without altering the sentence's meaning.
(\ldots) \\
\hline
(3) Which of the given options can replace the highlighted word in the given metaphorical sentence without altering the sentence's meaning?
(\ldots) \\
\hline
\textbf{M-word} \\
\hline
(1) Choose the word(s) that can replace the highlighted metaphorically used word in the given sentence without changing the meaning of the sentence.
(\ldots) \\
\hline
(2) Select words that can replace the highlighted metaphorically used word in the given sentence without altering the sentence's meaning.
(\ldots) \\
\hline
(3) Which of the given options can replace the highlighted metaphorically used word in the given sentence without altering the sentence's meaning?
(\ldots) \\
\hline
\end{tabular}
\caption{\label{tbl:prompts-word-judge}
Prompts for the word judgement task.}
\end{table}

\begin{table}[!t]
\small
\centering
\begin{tabular}{p{0.45\textwidth}}
\hline
\textbf{Implicit} \\
\hline
(1) Choose the correct paraphrase(s) for the given sentence. \\
Sentence: (a metaphor sample) \\
Option A: (a candidate paraphrase) \\
Option B: (another candidate paraphrase) \\
Option C: Both Option A and Option B \\
Option D: Neither Option A nor Option B \\
Correct answer: Option \\
\hline
(2) Select sentences that paraphrase the given sentence.
(\ldots) \\
\hline
(3) Select sentences that are semantically equivalent to the following sentence.
(\ldots) \\
\hline
\textbf{M-sent} \\
\hline
(1) Choose the correct paraphrase(s) for the given metaphorical sentence.
(\ldots) \\
\hline
(2) Select sentences that paraphrase the given metaphorical sentence.
(\ldots) \\
\hline
(3) Select sentences that are semantically equivalent to the following metaphorical sentence.
(\ldots) \\
\hline
\textbf{M-word} \\
\hline
(1) You are given a sentence where the highlighted word is metaphorically used. Choose the correct paraphrase(s) for the given sentence.
(\ldots) \\
\hline
(2) Given a sentence where the highlighted word is metaphorically used, select sentences that paraphrase this sentence.
(\ldots) \\
\hline
(3) Given a sentence where the highlighted word is metaphorically used, select sentences that are semantically equivalent to this sentence.
(\ldots) \\
\hline
\end{tabular}
\caption{\label{tbl:prompts-sent-judge}
Prompts for the sentence judgement task.}
\end{table}

\begin{table}[!t]
\centering
\small
\begin{tabular}{p{0.45\textwidth}}
\hline
(1) Paraphrase the given sentence by substituting the highlighted word with another word. The substitution should be a single word. \\
Sentence: No golden light $\ast$bathed$\ast$ the red brick of the house. \\
\textcolor{gray}{\texttt{llama:}} \\
Paraphrase: No golden light $\ast$[blank]$\ast$ the red brick of the house. \\
{[blank]} should be ``\_\_\_ \\
\textcolor{gray}{\texttt{gpt:}} \\
Paraphrase: No golden light $\ast$\_\_\_$\ast$ the red brick of the house. \\
\hline
(2) Use a single word to replace the highlighted word in the given sentence, so that the new sentence and the given sentence mean the same thing. \\
Sentence: (\ldots) \\
New sentence: (\ldots) \\
\hline
(3) Given a sentence with a highlighted word, replace this word with a different word to make a paraphrase. \\
Sentence: (\ldots) \\
Paraphrase: (\ldots) \\
\hline
\end{tabular}
\caption{\label{tbl:prompts-ppgen}
Prompts for the paraphrase generation task.
The blank (\_\_\_) denotes the place where models are asked to provide their answers:
The LLaMA models append answer after the left quotation mark (``) while GPT-3.5 inserts answer between the two asterisks ($\ast$).
The blank itself is not part of the prompts.}
\end{table}

We accessed the LLaMA models through Hugging Face; the queries used $\sim$880 GPU hours.
Our GPT-3.5 queries through the OpenAI API cost $\sim$255 USD.

We provide all the prompts used in this study:
three prompts for each condition of the paraphrase judgement tasks, including
word judgement (Table~\ref{tbl:prompts-word-judge}) and sentence judgement (Table~\ref{tbl:prompts-sent-judge});
and three prompts for the paraphrase generation task (Table~\ref{tbl:prompts-ppgen}).

%% file: body/a8_error.tex
\begin{table}[!t]
\centering
\small
\begin{tabular}{lrrr}
\hline
 & \texttt{llama-13b} & \texttt{llama-30b} &  \texttt{gpt-3.5} \\
\hline
\underline{A/B}    & & & \\
\textcolor{Green}{A/B} & 667 & 462 & 373 \\
\textcolor{red}{B/A} & 212 &  47 &  20 \\
\textcolor{red}{C}   & 193 & 563 & 641 \\
\textcolor{red}{D}   &   0 &   0 &  38 \\
\hline
\underline{C}      & & & \\
\textcolor{Green}{C}   &  10 &  32 &  38 \\
\textcolor{red}{A/B} &  35 &  13 &   5 \\
\textcolor{red}{D}   &   0 &   0 &   2 \\
\hline
\underline{D}      & & & \\
\textcolor{Green}{D}   &   0 &   0 & 101 \\
\textcolor{red}{A/B} & 241 &  33 &  52 \\
\textcolor{red}{C}   & 134 & 342 & 222 \\
\hline
\end{tabular}
\caption{\label{tbl:error_4opt}
Count for each combination of \underline{expected} answer and \textcolor{Green}{correct} or \textcolor{red}{incorrect} prediction when each model achieves their highest performance in the paraphrase judgement task.
A/B (A or B) means one of the two candidate paraphrases is expected or predicted as the correct answer.
The counts are based on the predictions of the models when they reach their respective highest accuracy in our experiments.}
\end{table}

\subsection{Paraphrase judgement}
\label{appx:error-ppjud}

Table~\ref{tbl:error_4opt} shows the number of each expectation-vs-prediction combination for each model when it achieved the highest accuracy score across all conditions and prompts:
The \textit{Metaphor-Word} condition of \textit{Word-judgement} for LLaMA-13B, using the third prompt (see Table~\ref{tbl:prompts-word-judge}); the \textit{Metaphot-Word} condition of \textit{Sentence-judgement} for LLaMA-30B, using the second prompt (see Table~\ref{tbl:prompts-sent-judge}); the \textit{Implicit} condition of \textit{Word-judgement} for GPT-3.5, using the third prompt.

\subsection{Factors associated with model performance}

Table~\ref{tbl:novelty-means-correct-vs-incorrect} summarises the novelty scores of the metaphor samples that receive correct versus incorrect answers from the models in the two paraphrase tasks.
Table~\ref{tbl:genre-acc} and \ref{tbl:pos-acc} show model accuracies in different genres and for different POS of the metaphorical word respectively.
Like in \ref{appx:error-ppjud},
the statistics are based on the best performance of each model.
In paraphrase generation, the LLaMA models achieve their respective best performance when given the first prompt (see Table~\ref{tbl:prompts-ppgen}); for GPT-3.5, it is the second prompt.

\begin{table}[!t]
\centering
\small
\begin{tabular}{l | c | c}
\hline
          & Judgement                 & Generation                \\
\hline
\texttt{llama-13b} & 0.07 / 0.07 & 0.07 / 0.06 \\
\texttt{llama-30b} & \textbf{0.05 / 0.08} & 0.06 / 0.06 \\ 
\texttt{gpt-3.5}   & 0.06 / 0.08 & \textbf{0.04 / 0.07} \\
\hline
\end{tabular}
\caption{\label{tbl:novelty-means-correct-vs-incorrect}
Mean novelty scores of metaphor samples that each model gives correct/incorrect answers when it achieves its respective highest performance in the paraphrase judgement and paraphrase generation tasks.
All standard deviations are $0.20 \pm 0.01$.
Boldface denotes that the difference between correct and incorrect answers is \textbf{statistically significant}.}
\end{table}

\begin{table}[!t]
\centering
\small
\begin{tabular}{lcccc}
\hline
 & \texttt{ACPROSE} & \texttt{NEWS} & \texttt{FICTION} & \texttt{CONVRSN} \\
\hline
\multicolumn{5}{l}{Judgement} \\
\texttt{llama-13b}
& .44 & .47 & .47 & - \\
\texttt{llama-30b}
& \textbf{.37} & .33 & \textbf{.24} & - \\
\texttt{gpt-3.5}
& .34 & .36 & .32 & - \\
\hline
\multicolumn{5}{l}{Generation} \\
\texttt{llama-13b}
& .15 & .17 & \textbf{.21} & \textbf{.13} \\
\texttt{llama-30b}
& .34 & .37 & \textbf{.37} & \textbf{.32} \\
\texttt{gpt-3.5}
& \textbf{.45} & .41 & \textbf{.40} & \textbf{.40} \\
\hline
\end{tabular}
\caption{\label{tbl:genre-acc}
Model accuracy in different genres when the models achieve their best performance in the paraphrase judgement and paraphrase generation tasks.
The metaphor samples for the paraphrase judgement task do not cover the conversation genre.
Boldface denotes \textbf{statistically significant} difference between the highest and lowest accuracies on the same row.
}
\end{table}

\begin{table}[!t]
\centering
\small
\begin{tabular}{lcccc}
\hline
 & N & V & A & R \\
\hline
\multicolumn{5}{l}{Judgement} \\
\texttt{llama-13b}
& .44 & .47 & .46 & \textcolor{lightgray}{.70} \\
\texttt{llama-30b}
& .34 & .32 & .30 & \textcolor{lightgray}{.30} \\
\texttt{gpt-3.5}
& \textbf{.38} & \textbf{.29} & .31 & \textcolor{lightgray}{.30} \\
\hline
\multicolumn{5}{l}{Generation} \\
\texttt{llama-13b}
& .18 & .16 & .15 & .13 \\
\texttt{llama-30b}
& \textbf{.37} & .36 & \textbf{.32} & .37 \\
\texttt{gpt-3.5}
& .44 & \textbf{.41} & \textbf{.40} & \textbf{.52} \\
\hline
\end{tabular}
\caption{\label{tbl:pos-acc}
Model accuracy per POS of the metaphorical word (\textbf{N}oun, \textbf{V}erb, \textbf{A}djective, and adve{\textbf{R}}b) when each model achieves its best performance in the paraphrase judgement and paraphrase generation tasks.
Boldface denotes \textbf{statistically significant} difference between the highest and lowest accuracies on the same row.
Accuracies for adverb metaphors in the paraphrase judgement task are \textcolor{lightgray}{disregarded} as the task only includes 10 adverb samples.}
\end{table}

%% file: main.bbl
\begin{thebibliography}{40}
\expandafter\ifx\csname natexlab\endcsname\relax\def\natexlab#1{#1}\fi

\bibitem[{Bang et~al.(2023)Bang, Cahyawijaya, Lee, Dai, Su, Wilie, Lovenia, Ji, Yu, Chung, Do, Xu, and Fung}]{bang-etal-2023-multitask}
Yejin Bang, Samuel Cahyawijaya, Nayeon Lee, Wenliang Dai, Dan Su, Bryan Wilie, Holy Lovenia, Ziwei Ji, Tiezheng Yu, Willy Chung, Quyet~V. Do, Yan Xu, and Pascale Fung. 2023.
\newblock \href {https://aclanthology.org/2023.ijcnlp-main.45} {A multitask, multilingual, multimodal evaluation of {C}hat{GPT} on reasoning, hallucination, and interactivity}.
\newblock In \emph{Proceedings of the 13th International Joint Conference on Natural Language Processing and the 3rd Conference of the Asia-Pacific Chapter of the Association for Computational Linguistics (Volume 1: Long Papers)}, pages 675--718, Nusa Dua, Bali. Association for Computational Linguistics.

\bibitem[{Bizzoni and Lappin(2018)}]{bizzoni-lappin-2018-predicting}
Yuri Bizzoni and Shalom Lappin. 2018.
\newblock \href {https://doi.org/10.18653/v1/W18-0906} {Predicting human metaphor paraphrase judgments with deep neural networks}.
\newblock In \emph{Proceedings of the Workshop on Figurative Language Processing}, pages 45--55, New Orleans, Louisiana. Association for Computational Linguistics.

\bibitem[{Brown et~al.(2020)Brown, Mann, Ryder, Subbiah, Kaplan, Dhariwal, Neelakantan, Shyam, Sastry, Askell, Agarwal, Herbert-Voss, Krueger, Henighan, Child, Ramesh, Ziegler, Wu, Winter, Hesse, Chen, Sigler, Litwin, Gray, Chess, Clark, Berner, McCandlish, Radford, Sutskever, and Amodei}]{brown-etal-2020-gpt3}
Tom Brown, Benjamin Mann, Nick Ryder, Melanie Subbiah, Jared~D Kaplan, Prafulla Dhariwal, Arvind Neelakantan, Pranav Shyam, Girish Sastry, Amanda Askell, Sandhini Agarwal, Ariel Herbert-Voss, Gretchen Krueger, Tom Henighan, Rewon Child, Aditya Ramesh, Daniel Ziegler, Jeffrey Wu, Clemens Winter, Chris Hesse, Mark Chen, Eric Sigler, Mateusz Litwin, Scott Gray, Benjamin Chess, Jack Clark, Christopher Berner, Sam McCandlish, Alec Radford, Ilya Sutskever, and Dario Amodei. 2020.
\newblock \href {https://proceedings.neurips.cc/paper_files/paper/2020/file/1457c0d6bfcb4967418bfb8ac142f64a-Paper.pdf} {Language models are few-shot learners}.
\newblock In \emph{Advances in Neural Information Processing Systems}, volume~33, pages 1877--1901. Curran Associates, Inc.

\bibitem[{Chakrabarty et~al.(2022{\natexlab{a}})Chakrabarty, Choi, and Shwartz}]{chakrabarty-etal-2022-rocket}
Tuhin Chakrabarty, Yejin Choi, and Vered Shwartz. 2022{\natexlab{a}}.
\newblock \href {https://doi.org/10.1162/tacl_a_00478} {It{'}s not rocket science: Interpreting figurative language in narratives}.
\newblock \emph{Transactions of the Association for Computational Linguistics}, 10:589--606.

\bibitem[{Chakrabarty et~al.(2022{\natexlab{b}})Chakrabarty, Saakyan, Ghosh, and Muresan}]{chakrabarty-etal-2022-flute}
Tuhin Chakrabarty, Arkadiy Saakyan, Debanjan Ghosh, and Smaranda Muresan. 2022{\natexlab{b}}.
\newblock \href {https://doi.org/10.18653/v1/2022.emnlp-main.481} {{FLUTE}: Figurative language understanding through textual explanations}.
\newblock In \emph{Proceedings of the 2022 Conference on Empirical Methods in Natural Language Processing}, pages 7139--7159, Abu Dhabi, United Arab Emirates. Association for Computational Linguistics.

\bibitem[{Choi et~al.(2021)Choi, Lee, Choi, Park, Lee, Lee, and Lee}]{choi-etal-2021-melbert}
Minjin Choi, Sunkyung Lee, Eunseong Choi, Heesoo Park, Junhyuk Lee, Dongwon Lee, and Jongwuk Lee. 2021.
\newblock \href {https://doi.org/10.18653/v1/2021.naacl-main.141} {{M}el{BERT}: Metaphor detection via contextualized late interaction using metaphorical identification theories}.
\newblock In \emph{Proceedings of the 2021 Conference of the North American Chapter of the Association for Computational Linguistics: Human Language Technologies}, pages 1763--1773, Online. Association for Computational Linguistics.

\bibitem[{Com{\textcommabelow{s}}a et~al.(2022)Com{\textcommabelow{s}}a, Eisenschlos, and Narayanan}]{comsa-etal-2022-miqa}
Iulia Com{\textcommabelow{s}}a, Julian Eisenschlos, and Srini Narayanan. 2022.
\newblock \href {https://aclanthology.org/2022.aacl-short.46} {{M}i{QA}: A benchmark for inference on metaphorical questions}.
\newblock In \emph{Proceedings of the 2nd Conference of the Asia-Pacific Chapter of the Association for Computational Linguistics and the 12th International Joint Conference on Natural Language Processing (Volume 2: Short Papers)}, pages 373--381, Online only. Association for Computational Linguistics.

\bibitem[{Czinczoll et~al.(2022)Czinczoll, Yannakoudakis, Mishra, and Shutova}]{czinczoll-etal-2022-scientific}
Tamara Czinczoll, Helen Yannakoudakis, Pushkar Mishra, and Ekaterina Shutova. 2022.
\newblock \href {https://doi.org/10.18653/v1/2022.findings-emnlp.153} {Scientific and creative analogies in pretrained language models}.
\newblock In \emph{Findings of the Association for Computational Linguistics: EMNLP 2022}, pages 2094--2100, Abu Dhabi, United Arab Emirates. Association for Computational Linguistics.

\bibitem[{Devlin et~al.(2019)Devlin, Chang, Lee, and Toutanova}]{devlin-etal-2019-bert}
Jacob Devlin, Ming-Wei Chang, Kenton Lee, and Kristina Toutanova. 2019.
\newblock \href {https://doi.org/10.18653/v1/N19-1423} {{BERT}: Pre-training of deep bidirectional transformers for language understanding}.
\newblock In \emph{Proceedings of the 2019 Conference of the North {A}merican Chapter of the Association for Computational Linguistics: Human Language Technologies, Volume 1 (Long and Short Papers)}, pages 4171--4186, Minneapolis, Minnesota. Association for Computational Linguistics.

\bibitem[{Do~Dinh et~al.(2018)Do~Dinh, Wieland, and Gurevych}]{do-dinh-etal-2018-weeding}
Erik-L{\^a}n Do~Dinh, Hannah Wieland, and Iryna Gurevych. 2018.
\newblock \href {https://doi.org/10.18653/v1/D18-1171} {Weeding out conventionalized metaphors: A corpus of novel metaphor annotations}.
\newblock In \emph{Proceedings of the 2018 Conference on Empirical Methods in Natural Language Processing}, pages 1412--1424, Brussels, Belgium. Association for Computational Linguistics.

\bibitem[{Gentner and Markman(1997)}]{gentner-markman-1997-structure}
D.~Gentner and A.~B Markman. 1997.
\newblock \href {https://doi.org/10.1037/0003-066X.52.1.45} {Structure mapping in analogy and similarity}.
\newblock \emph{American Psychologist}, 52(1):45--56.

\bibitem[{Grady et~al.(1999)Grady, Oakley, and Coulson}]{grady-etal-1999-blending}
Joseph Grady, Todd Oakley, and Seana Coulson. 1999.
\newblock Blending and metaphor.
\newblock \emph{AMSTERDAM STUDIES IN THE THEORY AND HISTORY OF LINGUISTIC SCIENCE SERIES 4}, pages 101--124.

\bibitem[{Hendrycks et~al.(2021)Hendrycks, Burns, Basart, Zou, Mazeika, Song, and Steinhardt}]{hendrycks-etal-2021-measuring}
Dan Hendrycks, Collin Burns, Steven Basart, Andy Zou, Mantas Mazeika, Dawn Song, and Jacob Steinhardt. 2021.
\newblock \href {http://arxiv.org/abs/2009.03300} {Measuring massive multitask language understanding}.

\bibitem[{Joseph et~al.(2023)Joseph, Liu, Ng, See, and Rai}]{joseph-etal-2023-newsmet}
Rohan Joseph, Timothy Liu, Aik~Beng Ng, Simon See, and Sunny Rai. 2023.
\newblock \href {https://doi.org/10.18653/v1/2023.findings-acl.641} {{N}ews{M}et : A {`}do it all{'} dataset of contemporary metaphors in news headlines}.
\newblock In \emph{Findings of the Association for Computational Linguistics: ACL 2023}, pages 10090--10104, Toronto, Canada. Association for Computational Linguistics.

\bibitem[{Kocoń et~al.(2023)Kocoń, Cichecki, Kaszyca, Kochanek, Szydło, Baran, Bielaniewicz, Gruza, Janz, Kanclerz, Kocoń, Koptyra, Mieleszczenko-Kowszewicz, Miłkowski, Oleksy, Piasecki, Łukasz Radliński, Wojtasik, Woźniak, and Kazienko}]{kocon-etal-2023-chatgpt}
Jan Kocoń, Igor Cichecki, Oliwier Kaszyca, Mateusz Kochanek, Dominika Szydło, Joanna Baran, Julita Bielaniewicz, Marcin Gruza, Arkadiusz Janz, Kamil Kanclerz, Anna Kocoń, Bartłomiej Koptyra, Wiktoria Mieleszczenko-Kowszewicz, Piotr Miłkowski, Marcin Oleksy, Maciej Piasecki, Łukasz Radliński, Konrad Wojtasik, Stanisław Woźniak, and Przemysław Kazienko. 2023.
\newblock \href {https://doi.org/https://doi.org/10.1016/j.inffus.2023.101861} {Chatgpt: Jack of all trades, master of none}.
\newblock \emph{Information Fusion}, page 101861.

\bibitem[{Lakoff(2014)}]{lakoff-2014-mapping}
George Lakoff. 2014.
\newblock \href {https://doi.org/10.3389/fnhum.2014.00958} {Mapping the brain's metaphor circuitry: metaphorical thought in everyday reason}.
\newblock \emph{Frontiers in Human Neuroscience}, 8.

\bibitem[{Lakoff and Johnson(1980)}]{lakoff-johnson-1980-metaphors}
George Lakoff and Mark Johnson. 1980.
\newblock \emph{Metaphors we live by}.
\newblock University of Chicago Press.

\bibitem[{Lan et~al.(2020)Lan, Chen, Goodman, Gimpel, Sharma, and Soricut}]{lan-etal-2020-albert}
Zhenzhong Lan, Mingda Chen, Sebastian Goodman, Kevin Gimpel, Piyush Sharma, and Radu Soricut. 2020.
\newblock \href {https://openreview.net/forum?id=H1eA7AEtvS} {Albert: A lite bert for self-supervised learning of language representations}.
\newblock In \emph{International Conference on Learning Representations}.

\bibitem[{Leong et~al.(2020)Leong, Beigman~Klebanov, Hamill, Stemle, Ubale, and Chen}]{leong-etal-2020-report}
Chee Wee~(Ben) Leong, Beata Beigman~Klebanov, Chris Hamill, Egon Stemle, Rutuja Ubale, and Xianyang Chen. 2020.
\newblock \href {https://doi.org/10.18653/v1/2020.figlang-1.3} {A report on the 2020 {VUA} and {TOEFL} metaphor detection shared task}.
\newblock In \emph{Proceedings of the Second Workshop on Figurative Language Processing}, pages 18--29, Online. Association for Computational Linguistics.

\bibitem[{Leong et~al.(2018)Leong, Beigman~Klebanov, and Shutova}]{leong-etal-2018-report}
Chee Wee~(Ben) Leong, Beata Beigman~Klebanov, and Ekaterina Shutova. 2018.
\newblock \href {https://doi.org/10.18653/v1/W18-0907} {A report on the 2018 {VUA} metaphor detection shared task}.
\newblock In \emph{Proceedings of the Workshop on Figurative Language Processing}, pages 56--66, New Orleans, Louisiana. Association for Computational Linguistics.

\bibitem[{Li et~al.(2023)Li, Wang, Lin, Guerin, and Barrault}]{li-etal-2023-framebert}
Yucheng Li, Shun Wang, Chenghua Lin, Frank Guerin, and Loic Barrault. 2023.
\newblock \href {https://doi.org/10.18653/v1/2023.eacl-main.114} {{F}rame{BERT}: Conceptual metaphor detection with frame embedding learning}.
\newblock In \emph{Proceedings of the 17th Conference of the European Chapter of the Association for Computational Linguistics}, pages 1558--1563, Dubrovnik, Croatia. Association for Computational Linguistics.

\bibitem[{Liang et~al.(2022)Liang, Bommasani, Lee, Tsipras, Soylu, Yasunaga, Zhang, Narayanan, Wu, Kumar, Newman, Yuan, Yan, Zhang, Cosgrove, Manning, Ré, Acosta-Navas, Hudson, Zelikman, Durmus, Ladhak, Rong, Ren, Yao, Wang, Santhanam, Orr, Zheng, Yuksekgonul, Suzgun, Kim, Guha, Chatterji, Khattab, Henderson, Huang, Chi, Xie, Santurkar, Ganguli, Hashimoto, Icard, Zhang, Chaudhary, Wang, Li, Mai, Zhang, and Koreeda}]{liang-etal-2022-holistic}
Percy Liang, Rishi Bommasani, Tony Lee, Dimitris Tsipras, Dilara Soylu, Michihiro Yasunaga, Yian Zhang, Deepak Narayanan, Yuhuai Wu, Ananya Kumar, Benjamin Newman, Binhang Yuan, Bobby Yan, Ce~Zhang, Christian Cosgrove, Christopher~D. Manning, Christopher Ré, Diana Acosta-Navas, Drew~A. Hudson, Eric Zelikman, Esin Durmus, Faisal Ladhak, Frieda Rong, Hongyu Ren, Huaxiu Yao, Jue Wang, Keshav Santhanam, Laurel Orr, Lucia Zheng, Mert Yuksekgonul, Mirac Suzgun, Nathan Kim, Neel Guha, Niladri Chatterji, Omar Khattab, Peter Henderson, Qian Huang, Ryan Chi, Sang~Michael Xie, Shibani Santurkar, Surya Ganguli, Tatsunori Hashimoto, Thomas Icard, Tianyi Zhang, Vishrav Chaudhary, William Wang, Xuechen Li, Yifan Mai, Yuhui Zhang, and Yuta Koreeda. 2022.
\newblock \href {http://arxiv.org/abs/2211.09110} {Holistic evaluation of language models}.

\bibitem[{Liu et~al.(2022)Liu, Cui, Zheng, and Neubig}]{liu-etal-2022-testing}
Emmy Liu, Chenxuan Cui, Kenneth Zheng, and Graham Neubig. 2022.
\newblock \href {https://doi.org/10.18653/v1/2022.naacl-main.330} {Testing the ability of language models to interpret figurative language}.
\newblock In \emph{Proceedings of the 2022 Conference of the North American Chapter of the Association for Computational Linguistics: Human Language Technologies}, pages 4437--4452, Seattle, United States. Association for Computational Linguistics.

\bibitem[{Mohammad et~al.(2016)Mohammad, Shutova, and Turney}]{mohammad-etal-2016-metaphor}
Saif Mohammad, Ekaterina Shutova, and Peter Turney. 2016.
\newblock \href {https://doi.org/10.18653/v1/S16-2003} {Metaphor as a medium for emotion: An empirical study}.
\newblock In \emph{Proceedings of the Fifth Joint Conference on Lexical and Computational Semantics}, pages 23--33, Berlin, Germany. Association for Computational Linguistics.

\bibitem[{Mohler et~al.(2016)Mohler, Brunson, Rink, and Tomlinson}]{mohler-etal-2016-introducing}
Michael Mohler, Mary Brunson, Bryan Rink, and Marc Tomlinson. 2016.
\newblock \href {https://aclanthology.org/L16-1668} {Introducing the {LCC} metaphor datasets}.
\newblock In \emph{Proceedings of the Tenth International Conference on Language Resources and Evaluation ({LREC}'16)}, pages 4221--4227, Portoro{\v{z}}, Slovenia. European Language Resources Association (ELRA).

\bibitem[{Pennington et~al.(2014)Pennington, Socher, and Manning}]{pennington-etal-2014-glove}
Jeffrey Pennington, Richard Socher, and Christopher Manning. 2014.
\newblock \href {https://doi.org/10.3115/v1/D14-1162} {{G}lo{V}e: Global vectors for word representation}.
\newblock In \emph{Proceedings of the 2014 Conference on Empirical Methods in Natural Language Processing ({EMNLP})}, pages 1532--1543, Doha, Qatar. Association for Computational Linguistics.

\bibitem[{Qin et~al.(2023)Qin, Zhang, Zhang, Chen, Yasunaga, and Yang}]{qin-etal-2023-chatgpt}
Chengwei Qin, Aston Zhang, Zhuosheng Zhang, Jiaao Chen, Michihiro Yasunaga, and Diyi Yang. 2023.
\newblock \href {https://doi.org/10.18653/v1/2023.emnlp-main.85} {Is {C}hat{GPT} a general-purpose natural language processing task solver?}
\newblock In \emph{Proceedings of the 2023 Conference on Empirical Methods in Natural Language Processing}, pages 1339--1384, Singapore. Association for Computational Linguistics.

\bibitem[{Shutova(2010)}]{shutova-2010-automatic}
Ekaterina Shutova. 2010.
\newblock \href {https://aclanthology.org/N10-1147} {Automatic metaphor interpretation as a paraphrasing task}.
\newblock In \emph{Human Language Technologies: The 2010 Annual Conference of the North {A}merican Chapter of the Association for Computational Linguistics}, pages 1029--1037, Los Angeles, California. Association for Computational Linguistics.

\bibitem[{Srivastava et~al.(2022)Srivastava, Rastogi, Rao, Shoeb, Abid, Fisch, Brown, Santoro, Gupta, Garriga-Alonso, Kluska, Lewkowycz, Agarwal, Power, Ray, Warstadt, Kocurek, Safaya, Tazarv, Xiang, Parrish, Nie, Hussain, Askell, Dsouza, Slone, Rahane, Iyer, Andreassen, Madotto, Santilli, Stuhlmüller, Dai, La, Lampinen, Zou, Jiang, Chen, Vuong, Gupta, Gottardi, Norelli, Venkatesh, Gholamidavoodi, Tabassum, Menezes, Kirubarajan, Mullokandov, Sabharwal, Herrick, Efrat, Erdem, Karakaş, Roberts, Loe, Zoph, Bojanowski, Özyurt, Hedayatnia, Neyshabur, Inden, Stein, Ekmekci, Lin, Howald, Diao, Dour, Stinson, Argueta, Ramírez, Singh, Rathkopf, Meng, Baral, Wu, Callison-Burch, Waites, Voigt, Manning, Potts, Ramirez, Rivera, Siro, Raffel, Ashcraft, Garbacea, Sileo, Garrette, Hendrycks, Kilman, Roth, Freeman, Khashabi, Levy, González, Perszyk, Hernandez, Chen, Ippolito, Gilboa, Dohan, Drakard, Jurgens, Datta, Ganguli, Emelin, Kleyko, Yuret, Chen, Tam, Hupkes, Misra, Buzan, Mollo, Yang, Lee, Shutova, Cubuk, Segal,
  Hagerman, Barnes, Donoway, Pavlick, Rodola, Lam, Chu, Tang, Erdem, Chang, Chi, Dyer, Jerzak, Kim, Manyasi, Zheltonozhskii, Xia, Siar, Martínez-Plumed, Happé, Chollet, Rong, Mishra, Winata, de~Melo, Kruszewski, Parascandolo, Mariani, Wang, Jaimovitch-López, Betz, Gur-Ari, Galijasevic, Kim, Rashkin, Hajishirzi, Mehta, Bogar, Shevlin, Schütze, Yakura, Zhang, Wong, Ng, Noble, Jumelet, Geissinger, Kernion, Hilton, Lee, Fisac, Simon, Koppel, Zheng, Zou, Kocoń, Thompson, Kaplan, Radom, Sohl-Dickstein, Phang, Wei, Yosinski, Novikova, Bosscher, Marsh, Kim, Taal, Engel, Alabi, Xu, Song, Tang, Waweru, Burden, Miller, Balis, Berant, Frohberg, Rozen, Hernandez-Orallo, Boudeman, Jones, Tenenbaum, Rule, Chua, Kanclerz, Livescu, Krauth, Gopalakrishnan, Ignatyeva, Markert, Dhole, Gimpel, Omondi, Mathewson, Chiafullo, Shkaruta, Shridhar, McDonell, Richardson, Reynolds, Gao, Zhang, Dugan, Qin, Contreras-Ochando, Morency, Moschella, Lam, Noble, Schmidt, He, Colón, Metz, Şenel, Bosma, Sap, ter Hoeve, Farooqi, Faruqui,
  Mazeika, Baturan, Marelli, Maru, Quintana, Tolkiehn, Giulianelli, Lewis, Potthast, Leavitt, Hagen, Schubert, Baitemirova, Arnaud, McElrath, Yee, Cohen, Gu, Ivanitskiy, Starritt, Strube, Swędrowski, Bevilacqua, Yasunaga, Kale, Cain, Xu, Suzgun, Tiwari, Bansal, Aminnaseri, Geva, Gheini, T, Peng, Chi, Lee, Krakover, Cameron, Roberts, Doiron, Nangia, Deckers, Muennighoff, Keskar, Iyer, Constant, Fiedel, Wen, Zhang, Agha, Elbaghdadi, Levy, Evans, Casares, Doshi, Fung, Liang, Vicol, Alipoormolabashi, Liao, Liang, Chang, Eckersley, Htut, Hwang, Miłkowski, Patil, Pezeshkpour, Oli, Mei, Lyu, Chen, Banjade, Rudolph, Gabriel, Habacker, Delgado, Millière, Garg, Barnes, Saurous, Arakawa, Raymaekers, Frank, Sikand, Novak, Sitelew, LeBras, Liu, Jacobs, Zhang, Salakhutdinov, Chi, Lee, Stovall, Teehan, Yang, Singh, Mohammad, Anand, Dillavou, Shleifer, Wiseman, Gruetter, Bowman, Schoenholz, Han, Kwatra, Rous, Ghazarian, Ghosh, Casey, Bischoff, Gehrmann, Schuster, Sadeghi, Hamdan, Zhou, Srivastava, Shi, Singh, Asaadi, Gu,
  Pachchigar, Toshniwal, Upadhyay, Shyamolima, Debnath, Shakeri, Thormeyer, Melzi, Reddy, Makini, Lee, Torene, Hatwar, Dehaene, Divic, Ermon, Biderman, Lin, Prasad, Piantadosi, Shieber, Misherghi, Kiritchenko, Mishra, Linzen, Schuster, Li, Yu, Ali, Hashimoto, Wu, Desbordes, Rothschild, Phan, Wang, Nkinyili, Schick, Kornev, Telleen-Lawton, Tunduny, Gerstenberg, Chang, Neeraj, Khot, Shultz, Shaham, Misra, Demberg, Nyamai, Raunak, Ramasesh, Prabhu, Padmakumar, Srikumar, Fedus, Saunders, Zhang, Vossen, Ren, Tong, Zhao, Wu, Shen, Yaghoobzadeh, Lakretz, Song, Bahri, Choi, Yang, Hao, Chen, Belinkov, Hou, Hou, Bai, Seid, Zhao, Wang, Wang, Wang, and Wu}]{srivastava-etal-2022-imitation}
Aarohi Srivastava, Abhinav Rastogi, Abhishek Rao, Abu Awal~Md Shoeb, Abubakar Abid, Adam Fisch, Adam~R. Brown, Adam Santoro, Aditya Gupta, Adrià Garriga-Alonso, Agnieszka Kluska, Aitor Lewkowycz, Akshat Agarwal, Alethea Power, Alex Ray, Alex Warstadt, Alexander~W. Kocurek, Ali Safaya, Ali Tazarv, Alice Xiang, Alicia Parrish, Allen Nie, Aman Hussain, Amanda Askell, Amanda Dsouza, Ambrose Slone, Ameet Rahane, Anantharaman~S. Iyer, Anders Andreassen, Andrea Madotto, Andrea Santilli, Andreas Stuhlmüller, Andrew Dai, Andrew La, Andrew Lampinen, Andy Zou, Angela Jiang, Angelica Chen, Anh Vuong, Animesh Gupta, Anna Gottardi, Antonio Norelli, Anu Venkatesh, Arash Gholamidavoodi, Arfa Tabassum, Arul Menezes, Arun Kirubarajan, Asher Mullokandov, Ashish Sabharwal, Austin Herrick, Avia Efrat, Aykut Erdem, Ayla Karakaş, B.~Ryan Roberts, Bao~Sheng Loe, Barret Zoph, Bartłomiej Bojanowski, Batuhan Özyurt, Behnam Hedayatnia, Behnam Neyshabur, Benjamin Inden, Benno Stein, Berk Ekmekci, Bill~Yuchen Lin, Blake Howald,
  Cameron Diao, Cameron Dour, Catherine Stinson, Cedrick Argueta, César~Ferri Ramírez, Chandan Singh, Charles Rathkopf, Chenlin Meng, Chitta Baral, Chiyu Wu, Chris Callison-Burch, Chris Waites, Christian Voigt, Christopher~D. Manning, Christopher Potts, Cindy Ramirez, Clara~E. Rivera, Clemencia Siro, Colin Raffel, Courtney Ashcraft, Cristina Garbacea, Damien Sileo, Dan Garrette, Dan Hendrycks, Dan Kilman, Dan Roth, Daniel Freeman, Daniel Khashabi, Daniel Levy, Daniel~Moseguí González, Danielle Perszyk, Danny Hernandez, Danqi Chen, Daphne Ippolito, Dar Gilboa, David Dohan, David Drakard, David Jurgens, Debajyoti Datta, Deep Ganguli, Denis Emelin, Denis Kleyko, Deniz Yuret, Derek Chen, Derek Tam, Dieuwke Hupkes, Diganta Misra, Dilyar Buzan, Dimitri~Coelho Mollo, Diyi Yang, Dong-Ho Lee, Ekaterina Shutova, Ekin~Dogus Cubuk, Elad Segal, Eleanor Hagerman, Elizabeth Barnes, Elizabeth Donoway, Ellie Pavlick, Emanuele Rodola, Emma Lam, Eric Chu, Eric Tang, Erkut Erdem, Ernie Chang, Ethan~A. Chi, Ethan Dyer, Ethan
  Jerzak, Ethan Kim, Eunice~Engefu Manyasi, Evgenii Zheltonozhskii, Fanyue Xia, Fatemeh Siar, Fernando Martínez-Plumed, Francesca Happé, Francois Chollet, Frieda Rong, Gaurav Mishra, Genta~Indra Winata, Gerard de~Melo, Germán Kruszewski, Giambattista Parascandolo, Giorgio Mariani, Gloria Wang, Gonzalo Jaimovitch-López, Gregor Betz, Guy Gur-Ari, Hana Galijasevic, Hannah Kim, Hannah Rashkin, Hannaneh Hajishirzi, Harsh Mehta, Hayden Bogar, Henry Shevlin, Hinrich Schütze, Hiromu Yakura, Hongming Zhang, Hugh~Mee Wong, Ian Ng, Isaac Noble, Jaap Jumelet, Jack Geissinger, Jackson Kernion, Jacob Hilton, Jaehoon Lee, Jaime~Fernández Fisac, James~B. Simon, James Koppel, James Zheng, James Zou, Jan Kocoń, Jana Thompson, Jared Kaplan, Jarema Radom, Jascha Sohl-Dickstein, Jason Phang, Jason Wei, Jason Yosinski, Jekaterina Novikova, Jelle Bosscher, Jennifer Marsh, Jeremy Kim, Jeroen Taal, Jesse Engel, Jesujoba Alabi, Jiacheng Xu, Jiaming Song, Jillian Tang, Joan Waweru, John Burden, John Miller, John~U. Balis,
  Jonathan Berant, Jörg Frohberg, Jos Rozen, Jose Hernandez-Orallo, Joseph Boudeman, Joseph Jones, Joshua~B. Tenenbaum, Joshua~S. Rule, Joyce Chua, Kamil Kanclerz, Karen Livescu, Karl Krauth, Karthik Gopalakrishnan, Katerina Ignatyeva, Katja Markert, Kaustubh~D. Dhole, Kevin Gimpel, Kevin Omondi, Kory Mathewson, Kristen Chiafullo, Ksenia Shkaruta, Kumar Shridhar, Kyle McDonell, Kyle Richardson, Laria Reynolds, Leo Gao, Li~Zhang, Liam Dugan, Lianhui Qin, Lidia Contreras-Ochando, Louis-Philippe Morency, Luca Moschella, Lucas Lam, Lucy Noble, Ludwig Schmidt, Luheng He, Luis~Oliveros Colón, Luke Metz, Lütfi~Kerem Şenel, Maarten Bosma, Maarten Sap, Maartje ter Hoeve, Maheen Farooqi, Manaal Faruqui, Mantas Mazeika, Marco Baturan, Marco Marelli, Marco Maru, Maria Jose~Ramírez Quintana, Marie Tolkiehn, Mario Giulianelli, Martha Lewis, Martin Potthast, Matthew~L. Leavitt, Matthias Hagen, Mátyás Schubert, Medina~Orduna Baitemirova, Melody Arnaud, Melvin McElrath, Michael~A. Yee, Michael Cohen, Michael Gu,
  Michael Ivanitskiy, Michael Starritt, Michael Strube, Michał Swędrowski, Michele Bevilacqua, Michihiro Yasunaga, Mihir Kale, Mike Cain, Mimee Xu, Mirac Suzgun, Mo~Tiwari, Mohit Bansal, Moin Aminnaseri, Mor Geva, Mozhdeh Gheini, Mukund~Varma T, Nanyun Peng, Nathan Chi, Nayeon Lee, Neta Gur-Ari Krakover, Nicholas Cameron, Nicholas Roberts, Nick Doiron, Nikita Nangia, Niklas Deckers, Niklas Muennighoff, Nitish~Shirish Keskar, Niveditha~S. Iyer, Noah Constant, Noah Fiedel, Nuan Wen, Oliver Zhang, Omar Agha, Omar Elbaghdadi, Omer Levy, Owain Evans, Pablo Antonio~Moreno Casares, Parth Doshi, Pascale Fung, Paul~Pu Liang, Paul Vicol, Pegah Alipoormolabashi, Peiyuan Liao, Percy Liang, Peter Chang, Peter Eckersley, Phu~Mon Htut, Pinyu Hwang, Piotr Miłkowski, Piyush Patil, Pouya Pezeshkpour, Priti Oli, Qiaozhu Mei, Qing Lyu, Qinlang Chen, Rabin Banjade, Rachel~Etta Rudolph, Raefer Gabriel, Rahel Habacker, Ramón~Risco Delgado, Raphaël Millière, Rhythm Garg, Richard Barnes, Rif~A. Saurous, Riku Arakawa, Robbe
  Raymaekers, Robert Frank, Rohan Sikand, Roman Novak, Roman Sitelew, Ronan LeBras, Rosanne Liu, Rowan Jacobs, Rui Zhang, Ruslan Salakhutdinov, Ryan Chi, Ryan Lee, Ryan Stovall, Ryan Teehan, Rylan Yang, Sahib Singh, Saif~M. Mohammad, Sajant Anand, Sam Dillavou, Sam Shleifer, Sam Wiseman, Samuel Gruetter, Samuel~R. Bowman, Samuel~S. Schoenholz, Sanghyun Han, Sanjeev Kwatra, Sarah~A. Rous, Sarik Ghazarian, Sayan Ghosh, Sean Casey, Sebastian Bischoff, Sebastian Gehrmann, Sebastian Schuster, Sepideh Sadeghi, Shadi Hamdan, Sharon Zhou, Shashank Srivastava, Sherry Shi, Shikhar Singh, Shima Asaadi, Shixiang~Shane Gu, Shubh Pachchigar, Shubham Toshniwal, Shyam Upadhyay, Shyamolima, Debnath, Siamak Shakeri, Simon Thormeyer, Simone Melzi, Siva Reddy, Sneha~Priscilla Makini, Soo-Hwan Lee, Spencer Torene, Sriharsha Hatwar, Stanislas Dehaene, Stefan Divic, Stefano Ermon, Stella Biderman, Stephanie Lin, Stephen Prasad, Steven~T. Piantadosi, Stuart~M. Shieber, Summer Misherghi, Svetlana Kiritchenko, Swaroop Mishra, Tal
  Linzen, Tal Schuster, Tao Li, Tao Yu, Tariq Ali, Tatsu Hashimoto, Te-Lin Wu, Théo Desbordes, Theodore Rothschild, Thomas Phan, Tianle Wang, Tiberius Nkinyili, Timo Schick, Timofei Kornev, Timothy Telleen-Lawton, Titus Tunduny, Tobias Gerstenberg, Trenton Chang, Trishala Neeraj, Tushar Khot, Tyler Shultz, Uri Shaham, Vedant Misra, Vera Demberg, Victoria Nyamai, Vikas Raunak, Vinay Ramasesh, Vinay~Uday Prabhu, Vishakh Padmakumar, Vivek Srikumar, William Fedus, William Saunders, William Zhang, Wout Vossen, Xiang Ren, Xiaoyu Tong, Xinran Zhao, Xinyi Wu, Xudong Shen, Yadollah Yaghoobzadeh, Yair Lakretz, Yangqiu Song, Yasaman Bahri, Yejin Choi, Yichi Yang, Yiding Hao, Yifu Chen, Yonatan Belinkov, Yu~Hou, Yufang Hou, Yuntao Bai, Zachary Seid, Zhuoye Zhao, Zijian Wang, Zijie~J. Wang, Zirui Wang, and Ziyi Wu. 2022.
\newblock \href {http://arxiv.org/abs/2206.04615} {Beyond the imitation game: Quantifying and extrapolating the capabilities of language models}.

\bibitem[{Srivastava(2022)}]{srivastava-2022-poirot}
Harshvardhan Srivastava. 2022.
\newblock \href {https://doi.org/10.18653/v1/2022.cmcl-1.11} {Poirot at {CMCL} 2022 shared task: Zero shot crosslingual eye-tracking data prediction using multilingual transformer models}.
\newblock In \emph{Proceedings of the Workshop on Cognitive Modeling and Computational Linguistics}, pages 102--107, Dublin, Ireland. Association for Computational Linguistics.

\bibitem[{Steen et~al.(2010{\natexlab{a}})Steen, Dorst, Herrmann, Kaal, and Krennmayr}]{steen-etal-2010-usage}
Gerard~J. Steen, Aletta~G. Dorst, J.~Berenike Herrmann, Anna~A. Kaal, and Tina Krennmayr. 2010{\natexlab{a}}.
\newblock \href {https://doi.org/10.1515/cogl.2010.024} {Metaphor in usage}.
\newblock \emph{Cognitive Linguistics}, 21(4):765--796.

\bibitem[{Steen et~al.(2010{\natexlab{b}})Steen, Dorst, Herrmann, Kaal, Krennmayr, and Pasma}]{steen-etal-2010-mipvu}
Gerard~J. Steen, Aletta~G. Dorst, J.~Berenike Herrmann, Anna~A. Kaal, Tina Krennmayr, and Tryntje Pasma. 2010{\natexlab{b}}.
\newblock \emph{A Method for Linguistic Metaphor Identification: From {MIP} to {MIPVU}}.
\newblock John Benjamins.

\bibitem[{Stowe et~al.(2022)Stowe, Utama, and Gurevych}]{stowe-etal-2022-impli}
Kevin Stowe, Prasetya Utama, and Iryna Gurevych. 2022.
\newblock \href {https://doi.org/10.18653/v1/2022.acl-long.369} {{IMPLI}: Investigating {NLI} models{'} performance on figurative language}.
\newblock In \emph{Proceedings of the 60th Annual Meeting of the Association for Computational Linguistics (Volume 1: Long Papers)}, pages 5375--5388, Dublin, Ireland. Association for Computational Linguistics.

\bibitem[{Tong(2021)}]{tong-2021-metaphor}
Xiaoyu Tong. 2021.
\newblock \href {https://scripties.uba.uva.nl/download?fid=681664} {Metaphor paraphrasing and word-sense disambiguation: toward a new approach to automated metaphor processin}.
\newblock Master's thesis, Universitetit van Amsterdam, the Netherlands.

\bibitem[{Tong et~al.(2021)Tong, Shutova, and Lewis}]{tong-etal-2021-recent}
Xiaoyu Tong, Ekaterina Shutova, and Martha Lewis. 2021.
\newblock \href {https://doi.org/10.18653/v1/2021.naacl-main.372} {Recent advances in neural metaphor processing: A linguistic, cognitive and social perspective}.
\newblock In \emph{Proceedings of the 2021 Conference of the North American Chapter of the Association for Computational Linguistics: Human Language Technologies}, pages 4673--4686, Online. Association for Computational Linguistics.

\bibitem[{Touvron et~al.(2023)Touvron, Lavril, Izacard, Martinet, Lachaux, Lacroix, Rozière, Goyal, Hambro, Azhar, Rodriguez, Joulin, Grave, and Lample}]{touvron-etal-2023-llama}
Hugo Touvron, Thibaut Lavril, Gautier Izacard, Xavier Martinet, Marie-Anne Lachaux, Timothée Lacroix, Baptiste Rozière, Naman Goyal, Eric Hambro, Faisal Azhar, Aurelien Rodriguez, Armand Joulin, Edouard Grave, and Guillaume Lample. 2023.
\newblock \href {http://arxiv.org/abs/2302.13971} {Llama: Open and efficient foundation language models}.

\bibitem[{Webb et~al.(2023)Webb, Holyoak, and Lu}]{webb-etal-2023-emergent}
Taylor Webb, Keith~J. Holyoak, and Hongjing Lu. 2023.
\newblock \href {http://arxiv.org/abs/2212.09196} {Emergent analogical reasoning in large language models}.

\bibitem[{Ye et~al.(2023)Ye, Chen, Xu, Zu, Shao, Liu, Cui, Zhou, Gong, Shen, Zhou, Chen, Gui, Zhang, and Huang}]{ye-etal-2023-comprehensive}
Junjie Ye, Xuanting Chen, Nuo Xu, Can Zu, Zekai Shao, Shichun Liu, Yuhan Cui, Zeyang Zhou, Chao Gong, Yang Shen, Jie Zhou, Siming Chen, Tao Gui, Qi~Zhang, and Xuanjing Huang. 2023.
\newblock \href {http://arxiv.org/abs/2303.10420} {A comprehensive capability analysis of gpt-3 and gpt-3.5 series models}.

\bibitem[{Zhang and Liu(2022)}]{zhang-liu-2022-metaphor}
Shenglong Zhang and Ying Liu. 2022.
\newblock \href {https://aclanthology.org/2022.coling-1.364} {Metaphor detection via linguistics enhanced {S}iamese network}.
\newblock In \emph{Proceedings of the 29th International Conference on Computational Linguistics}, pages 4149--4159, Gyeongju, Republic of Korea. International Committee on Computational Linguistics.

\bibitem[{Zhong et~al.(2023)Zhong, Ding, Liu, Du, and Tao}]{zhong-etal-2023-chatgpt}
Qihuang Zhong, Liang Ding, Juhua Liu, Bo~Du, and Dacheng Tao. 2023.
\newblock \href {http://arxiv.org/abs/2302.10198} {Can chatgpt understand too? a comparative study on chatgpt and fine-tuned bert}.

\end{thebibliography}
